\documentclass[10pt,journal,compsoc,onecolumn]{IEEEtran}

\usepackage{cite}
\usepackage{xr}
\usepackage[pdftex]{graphicx}
\graphicspath{{../pdf/}{../jpeg/}}
\DeclareGraphicsExtensions{.pdf,.jpeg,.png}
\usepackage{array}
\usepackage{amsmath}
\usepackage{amssymb}
\usepackage{mdwmath}
\usepackage{mdwtab}
\usepackage{eqparbox}
\usepackage{url}
\usepackage{hyperref}
\usepackage[switch]{lineno}
\usepackage{xcolor}

\usepackage{tcolorbox}
\tcbset{standard jigsaw, opacityback=0.5}

\usepackage{graphicx}
\usepackage{multirow}
\graphicspath{{figures/}}

\usepackage{xspace}
\usepackage{subfig}
\usepackage{algorithm}
\usepackage[noend]{algpseudocode}
\usepackage{tikz}
\usepackage{booktabs}
\usepackage{gensymb}
\usepackage{tabularx}
\usepackage[numbers]{natbib}
\usepackage{titlesec}
\titleformat{\paragraph}
{\normalfont\small}{\theparagraph}{1em}{}
\titlespacing*{\paragraph}
{0pt}{3.25ex plus 1ex minus .2ex}{1.5ex plus .2ex}

\newcommand{\ours}{DeCaPH\xspace}
\MakeRobust\ours

\newcommand{\oursFull}{Decentralised, Collaborative, and Privacy-preserving Machine Learning for Multi-Hospital Data\xspace}
\newcommand{\oursFullBold}{\textbf{\underline{De}}centralised, \textbf{\underline{C}}ollaborative, \textbf{\underline{a}}nd \textbf{\underline{P}}rivacy-preserving Machine Learning for Multi-\textbf{\underline{H}}ospital Data\xspace}
\MakeRobust\oursFull

\DeclareRobustCommand\encircle[1]{\tikz[baseline=(char.base)]{\node[shape=circle,fill,inner sep=1pt] (char) {\textcolor{white}{#1}}}}

\newcommand{\Dataset}{\mathcal{D}}

\title{Decentralised, Collaborative, and Privacy-preserving Machine Learning for Multi-Hospital Data}
\author{Congyu Fang, Adam Dziedzic, Lin Zhang, Laura Oliva, Amol Verma, Fahad Razak, Nicolas Papernot*, Bo Wang* \IEEEcompsocitemizethanks{\IEEEcompsocthanksitem Congyu Fang is with the Department of Department of Computer Science, University of Toronto; Peter Munk Cardiac Centre, University Health Network; Vector Institute, Toronto, Canada\\
Adam Dziedzic is with CISPA Helmholtz Center for Information Security. Work done while the author is at Department of Electrical and Computer Engineering, University of Toronto; Vector Institute, Toronto, Canada\\
Lin Zhang is with Simon Fraser University. Work done while the author was at Peter Munk Cardiac Centre, University Health Network\\
Laura Oliva is with Peter Munk Cardiac Centre, University Health Network\\
Amol Verma and Fahad Razak are with St. Michael’s Hospital, Unity Health Toronto; Department of Medicine, University of Toronto; Institute of Health Policy, Management and Evaluation, University of Toronto\\
Nicolas Papernot is with Department of Electrical and Computer Engineering, University of Toronto; Department of Computer Science, University of Toronto; Vector Institute, Toronto, Canada\\
Bo Wang is with the Department of Laboratory Medicine and Pathobiology, Temerty Faculty of Medicine, University of Toronto; Department of Computer Science, University of Toronto; Peter Munk Cardiac Centre, University Health Network; Vector Institute, Toronto, Canada\\
\protect\\

\IEEEcompsocthanksitem Corresponding author: Bo Wang.
E-mail: bowang@vectorinstitute.ai; Nicolas Papernot. Email: nicolas.papernot@utoronto.ca}}

\date{Jan 31, 2024}

\begin{document}
\maketitle
\section*{Abstract}
\noindent \textbf{Background:}
Machine Learning (ML) has demonstrated its great potential on medical data analysis. 
Large datasets collected from diverse sources and settings are essential for ML models in healthcare to achieve better accuracy and generalizability. Sharing data across different healthcare institutions or jurisdictions is challenging because of complex and varying privacy and regulatory requirements. Hence, it is hard but crucial to allow multiple parties to collaboratively train an ML model leveraging the private datasets available at each party without the need for direct sharing of those datasets or compromising the privacy of the datasets through collaboration. 

\noindent \textbf{Methods:} In this paper, we address this challenge by proposing \textbf{De}centralized, \textbf{C}ollaborative, \textbf{a}nd \textbf{P}rivacy-preserving ML for Multi-\textbf{H}ospital Data (\ours). This framework offers the following key benefits: (1) it allows different parties to collaboratively train an ML model without transferring their private datasets (i.e., no data centralization); (2) it safeguards patients' privacy by limiting the potential privacy leakage arising from any contents shared across the parties during the training process; and (3) it facilitates the ML model training without relying on a centralized party/server. 

\noindent \textbf{Findings: } We demonstrate the generalizability and power of DeCaPH on three distinct tasks using real-world distributed medical datasets: patient mortality prediction using electronic health records, cell-type classification using single-cell human genomes, and pathology identification using chest radiology images. The ML models trained with \ours framework have less than 3.2\% drop in model performance comparing to those trained by the non-privacy-preserving collaborative framework. Meanwhile, the average vulnerability to privacy attacks of the models trained with \ours decreased by up to 16\%. In addition, models trained with our \ours framework achieve better performance than those models trained solely with the private datasets from individual parties without collaboration and those trained with the previous privacy-preserving collaborative training framework under the same privacy guarantee by up to 70\% and 18.2\% respectively.

\noindent \textbf{Interpretation: } We demonstrate that the ML models trained with \ours framework have an improved utility-privacy trade-off, showing \ours enables the models to have good performance while preserving the privacy of the training data points. 
In addition, the ML models trained with \ours framework in general outperform those trained solely with the private datasets from individual parties, showing that \ours enhances the model generalizability.

\noindent \textbf{Funding: }This work was supported by the Natural Sciences and Engineering Research Council of Canada (NSERC, RGPIN-2020-06189 and DGECR-2020-00294), Canadian Institute for Advanced Research (CIFAR) AI Catalyst Grants,  CIFAR AI Chair programs, Temerty Professor of AI Research and Education in Medicine, University of Toronto, Amazon, Apple, DARPA through the GARD project, Intel, Meta, the Ontario Early Researcher Award, and the Sloan Foundation. Resources used in preparing this research were provided, in part, by the Province of Ontario, the Government of Canada through CIFAR, and companies sponsoring the Vector Institute.

{\it Keywords}: Collaborative Machine Learning (ML), (Distributed) Differential Privacy, Decentralization, ML for healthcare. 
\newpage

\section*{Introduction}
\label{sec:intro}
Machine Learning (ML) models have shown great potential for medical data analysis~\cite{yu_artificial_2018, LITJENS2017survey}, such as medical imaging analysis~\cite{Varoquaux2022ml_medical_imaging}, genome interpretation~\cite{Libbrecht2015ml_genomics}, and clinical outcome prediction~\cite{Shamout2021_ml_clinical}. These advancements could potentially aid human experts in decision-making processes such as disease detection~\cite{esteva2017dermatologist}, annotation of pathogenic gene variants~\cite{quang_dann_2015}.  ML models typically benefit from large volumes of training data from diverse sources for improved generalizability, for example, in the study of histopathology, the datasets used by current studies do not include a sufficient number of laboratories to demonstrate generalizability~\cite{van_der_laak_deep_2021}. Ideally, aggregating the healthcare datasets from different hospitals and institutes and jointly training an ML model would achieve better model accuracy and generalizability~\cite{rieke_future_2020,pfitzner_federated_2021,ng_federated_2021,sheller_federated_2020}. 
However, healthcare data usually contains highly sensitive information; 
data sharing across multiple institutions can threaten patients' privacy and is often subject to complex privacy regulations~\cite{gdpr_medical} that differ across jurisdictions.
There is also a risk associated with the model weights revealing information about their private training datasets. To reason about privacy in this context, the current golden standard is differential privacy (DP)~\cite{Dwork2006calibratingnoise, Dwork2011afirmfoundation, Dwork2014algorithmicfoundationDP}. It offers a strong guarantee with no assumptions about the capability or goals of potential adversaries. It provides a theoretical upper bound (often known as the privacy budget, $\epsilon$) on the potential privacy leakage of a randomized algorithm that uses a dataset as its input. It represents how much information can leak about the training data in the worst case.

Numerous works have been conducted to develop collaborative ML training frameworks. Federated Learning (FL) is one of the earliest ~\cite{mcmahan2017fl}. It employs a central server to coordinate a set of clients (e.g., hospitals) to jointly train a model. At the training stage, each client locally computes their model updates on their private datasets. These updates are then sent to the central server for merging. To prevent the server from viewing the individual clients’ updates, the central server can perform the merging via employing secure aggregation (SecAgg)~\cite{Bonawitz2017SecAgg, Bell2020SecAggPlus}, a cryptographic approach to securely compute a summation over multiple parties' model updates while disclosing the clients' model updates to the server. Even though FL protects the confidentiality of the private datasets, the models trained with FL are not differentially private, meaning that it cannot formally guarantee the privacy of the data points used in training.

To protect data privacy at an individual data point level, Privacy-preserving Medical Image Analysis (PriMIA)~\cite{Kaissis2021primia} combines FL with differentially private stochastic gradient descent (DP-SGD)~\cite{Abadi2016dpsgd} and Secure Aggregation (SecAgg). However, similar to FL, PriMIA also requires a central server, which impedes the framework's scalability due to the computational overhead required for the server to aggregate the model updates and facilitate training. 

Eliminating a central party would enhance a collaborative ML protocol's flexibility and robustness, such as improved transparency during training and the avoidance of single-point failure. Addressing this limitation, Swarm Learning (SL)~\cite{Warnat-Herresthal2021swarmlearning}, on the other hand, is a decentralised FL approach. It employs blockchain technology to enable secure onboarding of participants. It also removes the central server by dynamically selecting the first party that completes training as the leader to facilitate the aggregation of model updates. Also, there are multiple concurrent works that combine blockchain technology with FL to achieve decentralisation of FL. One proposed approach involves having the miners/participants compete for the leader role to perform the aggregation~\citep{ma2022when_FL_meets_blockchain, Zhao2021PP_blockchain_fl}.  However, this could lead to the repeated selection of the same party (that has the most computation power) as the leader, defeating the purpose of rotating leadership roles. Most importantly, the employment of blockchain technology itself does not provide any privacy guarantees, making it not sufficient to protect the privacy of patients' information. In addition, given that hospitals must adhere to strict legislation and restrictions and with minimal chance of them being adversarial, having participants compete for the leadership role could be unnecessary. It also introduces additional computation to each participants for mining, which will slow down the entire model training process. 

Therefore, a new framework which allows different parties to securely collaborate while safeguarding the privacy of their private datasets (i.e., satisfying DP) is in demand. Based on our analysis of the existing frameworks and considerations on the needs for healthcare research as well as the sensitive nature of healthcare datasets, we identify the following key requirements for a secure ML training framework that enables collaboration among hospitals while preserving the privacy of each hospital's private datasets:
\begin{enumerate}
\item No transfer of the private datasets should occur (i.e., raw data shall not be revealed among the entities) such that the private datasets of each participant would remain confidential and decentralised. 
 
\item When sharing computed/derived information from private datasets among the participants (e.g., gradients, intermediate and final model weights), there should be a theoretical upper bound on the potential privacy leakage of private training data via these shared contents. 
\item The parties should be able to collaboratively train an aggregate model without the existence of a centralized party. 
\end{enumerate}

\begin{tcolorbox}[colback=purple!30]

\section*{Research in context}

\subsection*{Evidence before this study}
Previously proposed collaborative training frameworks either do not provide the correct level of privacy protection for individual patients or achieve the best utility-privacy trade-off under the specific requirements of hospitals. In addition, previous studies usually lack analysis on real-world datasets collected from multiple hospitals to demonstrate the capability of the frameworks. We searched PubMed, Nature, IEEE, NeurIPS, and ICLR for journals and conference articles, using the terms ``distributed training'', ``collaborative training'', ``federated learning'', ``privacy-preserving'', ``differential privacy'', ``distributed differential privacy'', ``global differential privacy'', ``healthcare'', and ``medical''. To the best of our knowledge, there are no previous studies that address all the required elements to enable collaborative healthcare research among hospitals to achieve the best privacy-utility trade-offs and conduct experiments using real-world datasets for multiple types of healthcare-related tasks/datasets.

\subsection*{Added value of this study}
We explicitly analysed the potential adversarial behaviours that may happen during the collaborations of hospitals and identified the required components to protect patient-level privacy to achieve the best privacy-utility trade-off for the trained models. We conducted extensive experiments using real-world cross-silo datasets on three tasks (clinical outcome prediction using electronic health records, cell classification using single-cell RNA transcriptomics, and pathology identification using chest radiology images). We showed the models trained with the proposed frameworks can provide privacy protection while having better performance than models trained without collaboration or previously proposed privacy-preserving frameworks. It demonstrates our framework is capable of supporting researchers to train models that generalize better to the broader population for various tasks. In addition, we conducted empirical privacy analysis, demonstrating the models trained with the proposed framework are much less vulnerable to privacy attacks. 

\subsection*{Implications of all the available evidence}
The proposed collaborative training
framework enables researchers to have access to a broader pool of data points to train more accurate and generalizable ML models while protecting the privacy of the patients. These models have the potential to enhance
the accessibility and affordability of healthcare services, offering valuable support to doctors in
areas like diagnosis and treatment recommendations, ultimately leading to improved patient care.

\end{tcolorbox}

To meet the aforementioned desired properties of a collaborative learning framework for healthcare research, we hereby propose \oursFullBold (\ours). It is a collaborative ML training framework that leverages randomized leader selection, secure aggregation, gradient clipping, and noising. In this framework, we eliminate the usage of a central server; all parties participating in \ours framework are referred to as participants (instead of clients). We evaluate the performance of \ours using three different healthcare-related tasks.

In specific, we contribute the following in this paper:
\begin{enumerate}
        \item We propose \ours, a collaborative ML training framework that ensures decentralisation and secure aggregation of participants' contributions. Notably, the models trained with \ours conform to Differential Privacy (DP), the gold standard for privacy in learning algorithms. 
    \item Our \ours framework offers theoretical DP guarantees under an \textbf{honest-but-curious} adversary model. This assumes participants will adhere to the protocol and not deliberately sabotage the training process, given our target users (hospitals and healthcare research institutes) are bounded by strong patient-centred ethical principles and subject to strict legislation. However, these participants might be interested in learning from the contributions of others, thereby justifying this threat model.
    \item We empirically evaluate \ours on three distinct tasks: predicting patient survival/mortality using electronic health records, classifying cell types from single-cell human genomes, and identifying pathologies from chest radiology images. These tasks demonstrate that \ours framework can effectively handle multiple modalities of healthcare-related data. 
    
    \item We conduct a membership inference attack~\cite{shokri2017membership,carlini2022membership} to empirically validate that the models trained with \ours framework are more robust against privacy attacks than those trained using existing collaborative learning frameworks that lack privacy guarantees, such as FL and SL. 
\end{enumerate}

The aim for various parties to collaborate is to utilise larger and more diverse datasets to improve ML models. Thus, the primary evaluation metric for the collaborative training framework is its ability to train an aggregate model that outperforms models trained only on the private datasets available at each silo.  The framework must also ensure that the collaboration process is privacy-preserving, i.e., the privacy leakage during and after collaboration is upper-bounded by a theoretical threshold. Consequently, an effective privacy-preserving collaborative framework should train models with good utility, while demonstrating superior robustness to privacy attacks than models trained without privacy-preserving mechanisms.

For the rest of this paper, we first introduce our proposed framework, \ours, followed by an overview of the three healthcare tasks used to evaluate \ours and their corresponding evaluation metrics. Subsequently, we present the Results section which describes dataset characteristics, sizes, and the machine learning models trained using \ours; models trained with \ours are compared with those trained with previous frameworks across various performance measures to demonstrate models trained \ours have improved privacy-utility trade-offs. In addition to performance assessment, we conduct an ablation study to demonstrate the significance of integrating privacy-preserving techniques into collaborative training frameworks. Specifically, we evaluate the models trained with \ours against models trained without DP in terms of their robustness to Membership Inference Attacks~\cite{shokri2017membership,carlini2022membership}, a common method used to empirically audit the privacy guarantee of a model. We then provide a Conclusion and Discussion section to summarise the contributions of the paper and discuss the potential future directions. Lastly, we present the Data Sharing section as well as the Detailed Methods section, providing necessary details about data preprocessing, framework pipelines, privacy analysis techniques employed by \ours, computations, algorithms, and the evaluation metrics used to assess the performance of the trained machine learning models. Additionally, we include more information on existing frameworks, empirical privacy analyses, experimental setups, results, dataset collection, and an analysis of framework communication costs in the Supplementary Materials.

\section*{Methods}
\label{sec:method}
\subsection*{Framework design (Overview): \ours }
\label{subsec:overview}

\begin{figure}[H]
\centering\hspace{-5mm}
\includegraphics[width=\linewidth]{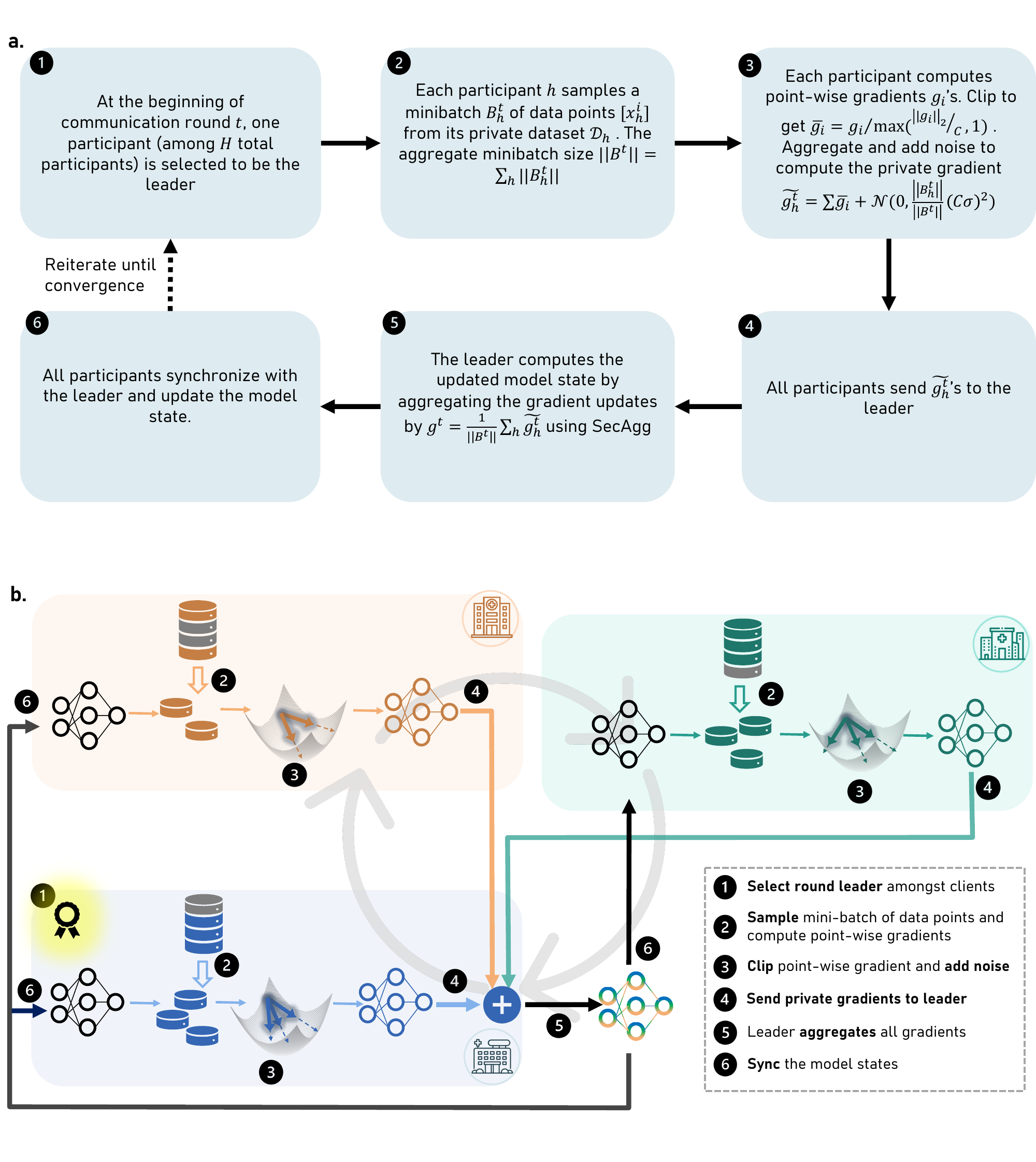}
\caption{ \textbf{An overview of \ours learning framework.} (\textbf{a}), flowchart of the steps for one iteration of training with \ours. At each communication round, \encircle{1} a leader is first selected to perform the \underline{aggregation of the participants' model weights}; \encircle{2} each hospital locally randomly samples a mini-batch of data points and computes their point-wise gradients; \encircle{3} each hospital locally clips the point-wise gradient vectors and adds a calibrated Gaussian Noise; \encircle{4} all participating hospitals send their local gradients to the leader; \encircle{5} the leader aggregates the gradients from all hospitals using SecAgg and outputs \underline{an updated model} that is \underline{differentially private}; \encircle{6} all participating hospitals synchronize their model state with the leader. \underline{Reiterate these steps until convergence.}  (\textbf{b}), visualization of one training iteration of \ours with three participating hospitals.}

\label{fig:framework_overview}
\end{figure}

The \ours decentralised collaborative framework is outlined in Figure~\ref{fig:framework_overview}. {\ours uses sampled Gaussian Mechanisms~\cite{mironov2019renyi}\footnote{\cite{mironov2019renyi} is a pre-print. } to train models with DP, which includes a few steps: random subsampling of training data points, bounding the contribution of each data points, and addition of Gaussian noise. These steps are completed by step 2 and 3 in \ours. Specifically,} before the training starts, all participating hospitals will communicate the sizes of their private datasets to determine a mini-batch sampling rate, $p$. This sampling rate will be used for the rest of the training. At the beginning of each communication round, a leader is randomly selected. The role of the leader is to aggregate the participants' model updates and facilitate the training process. Then all the participants randomly sample a mini-batch of data points based on the sampling probability, $p$, and compute the point-wise gradient updates. Each participant locally clips the point-wise gradient vectors and adds a calibrated Gaussian Noise to the clipped gradient vectors. These clipped and noised gradient updates are then sent to the leader and the leader will merge those updates using Secure Aggregation (SecAgg) to output an updated model state. The usage of 
mini-batch subsampling, gradient clipping, and noising in \ours framework offers distributed (or global) DP (DDP) guarantees under an honest-but-curious threat model. To mathematically quantify the privacy guarantee of the training algorithm, we compute a privacy budget, $\epsilon$, which represents the worst-case information leakage that can happen. The final step of each communication round is for all the participants to synchronize their model states with the leader. Then a new leader is selected for the next communication round. These steps are repeated until the model converges or a predetermined privacy budget $\epsilon$ is reached. 

{
Note that in \ours, the leader that facilitates the training process is selected randomly for each round. This strategy is enough for the intended application scenario because the participants of the framework are hospitals, who are expected to honestly adhere to the protocol. A more formal discussion about such an honest-but-curious threat model is provided in the Detailed Framework and Study Design subsection. Hence, more complex leader selection strategies that can prevent malicious participants are not necessary. The purpose of random selection of leader is to rotate the role of facilitating the process to all participants to improve scalability, avoid single-point of failure, distribute the additional computational costs of the leader, and improve transparency. Techniques like distributed ledger, cryptography, smart contracts are complementary to our framework. That is one can integrate such techniques (e.g., blockchain) with our framework to facilitate the onboarding of participants, logging the training process etc., hence we will not discuss it further for purpose of this paper. }

{In addition, with the honest-but-curious threat model,} the leader of each round will exclusively have access to the aggregated model weights generated from the SecAgg algorithm.  The clipping and noising procedures of \ours will make sure the aggregated model weights satisfy DDP to provide privacy protection to all patients of all participants. In contrast, previous frameworks like PriMIA provide local DP. Local DP is necessary if the aggregator of each round can not be trusted to follow the protocol so that the participants' model updates must be privacy-protected before submitting to the aggregator. However, this is not a concern in our threat model, as the aggregator cannot access individual participant's model updates. Achieving local DP also involves adding more noise and causing more performance degradation in comparison to achieving DDP as proposed by \ours. Hence \ours is able to achieve the best privacy-utility trade-off. More technical details are provided in the {Detailed Framework and Study Design subsection}. 

\subsection*{Study design (Overview)}
We will assess the performance and demonstrate the capability of \ours framework on three different tasks with three real-world cross-silo healthcare datasets: electronic health records (EHR), single-cell RNA-seq of human pancreas, and human chest radiology images. {After necessary filtering, the EHR, presented as a tabular dataset, contains 40,114 unique health records of patients who are discharged or dead within 24 hours of admission collected from eight hospitals in Ontario, Canada. It contains both numerical and categorical features (436 input features in total).} The number of input features has a relatively low dimension, but the data requires a lot of cleanup and standardization. Single-cell RNA-seq data {used in the analysis} also comes as tabular datasets. {It is collected from five studies which contains 10,548 cells. In comparison to EHR,} it is more structured but it has a lot more input features {(15,558 input features). Each input feature represents the counts of each gene expression.} {The last datasets used in the analysis are the} human chest radiology datasets, {which} are medical image datasets. {Those datasets are acquired from three studies, containing 267,953 images in total. }These datasets are acquired from multiple hospitals and studies, making them ideal for demonstrating the ability of the collaborative framework to handle the imbalance and heterogeneity of real-world datasets. {More details about the three datasets are provided in Figures~\ref{fig:results_case_a},~\ref{fig:results_case_b}, and~\ref{fig:results_case_c}, and Preprocessing of the Detailed Framework and Study Design subsection, and the Data Collection section of the Supplementary Materials. }

For each of the case studies, we will present the proportions of each participant’s private dataset size at each silo and the balance of classes in each private dataset. In addition, we also compare the performance of the models trained by \ours framework with those trained by conventional FL~\cite{mcmahan2017fl}, PriMIA~\cite{Kaissis2021primia}, as well as the models trained locally at each silo using only the private data available at this silo. Note that SL (and other blockchain equipped FL frameworks) can be considered equivalent to FL when comparing model performance and DP guarantee of their trained models, since SL is a decentralised implementation of FL. The key distinction is that SL does not use a central server. Both frameworks utilise the same training algorithm and offer the same DP guarantee (i.e., no DP guarantee). Therefore, there is no need to include a separate comparison for model performance or model robustness to privacy attacks of SL-trained models. 
{Also, since FL and its variants are not privacy-preserving, models trained with FL represent the best model performance/utility that can be reached without considering privacy. Hence, in order to demonstrate the privacy-utility trade-off of \ours, performance of models trained with \ours is compared to that of FL to calculate the percentage drop in model's performance in exchange for a reasonable privacy guarantee; }{in addition to FL, PriMIA is included to demonstrate the privacy-utility trade-off of \ours since PriMIA is the only collaborative framework (to the best of our knowledge) that protect the patient-level privacy as required for healthcare related tasks. Hence, it is used for comparison with \ours to show if \ours can improve the performance of the model when trained with the same privacy guarantee. More existing collaborative frameworks are discussed in Supplementary materials and we explain why they are not feasible in the context of hospital collaborations.
} 

{For all the experiments, each study/hospital is treated as one participant in the framework possessing their own private dataset. Each participant has access only to its private dataset and collaboratively train an ML model following the steps outlined in Figure~\ref{fig:framework_overview}. More details about the hyperparameters and algorithms used for training the ML model are presented in Computation and algorithms of the {Detailed Framework and Study Design subsection} section and Experimental Setup of the Supplementary Materials.}

{Different metrics are used to evaluate for the performance of the ML models trained for each different tasks. For example, metrics like PPV and NPV are used when prediction the mortality of patients, whereas weighted precision and recall are used for cell type classification. The Area under the Receiver (AUROC) is used to evaluate the performance of models on pathology identification task These evaluation metrics are specific to the tasks. More details are provided in Evaluation Metrics of Detailed Framework and Study Design subsection. }

\subsection*{Detailed Framework and Study Design}
\label{sec:methods} 

\subsubsection*{Preprocessing}
\label{ssec:preprocessing}

\paragraph*{GEMINI} GEMINI data are collected from hospital information systems and aggregated to a central repository. Access to data can be obtained upon reasonable request and in line with local ethics and privacy protocols, via \url{https://www.geminimedicine.ca/}. A rigorous process for data quality control is applied, including computation and manual data validation~\cite{Verma2017gemini,Verma2021_gemini}. Hospital administrative data are standardized by hospitals for reporting to the Canadian Institute for Health Information. Clinical data are extracted in various formats from different hospital systems and standardized centrally by the GEMINI team in alignment with the OMOP common data model. 

The cohort for this study includes inpatients admitted to General Internal Medicine (GIM). The datasets contain both categorical features like triage level, as well as numerical values like age, and measures from the lab tests. Categorical features are one-hot encoded and numerical features are normalized to have a mean of 0 and standard deviation of 1. More details about data collection, inclusion/exclusion criteria, and the features used for this study is provided in the Data Collection section in the Supplementary materials.

\paragraph*{Single Cell Human Pancreas} 
The detailed preprocessing steps are described by~\cite{Wang2022ocat} and we use the preprocessed version available at \url{https://data.wanglab.ml/OCAT/Pancreas.zip}. In these datasets, each entry $r_{ij}$ represents the count of gene expression $j$ for cell $i$. We apply log transformation to each of the entries, i.e., $r_{ij} \leftarrow log_{10} (r_{ij}+1)$. The four common cell types (alpha, beta, gamma, and delta) are one-hot encoded and used as the classification labels.

\paragraph*{Chest Radiology}
We use chest X-Ray dataset from the National Institute of Health (NIH)~\cite{Wang2017nih-chestx-ray8}, PadChest (PC)~\cite{bustos_padchest_2020}, and CheXpert (CheX)~\cite{irvin2019chexpert}. These three studies form the participants 1 to 3, in that order, in the third case study. We also use MIMIC-CXR~\cite{Johnson2019mimic,Johnson2019mimic-physionet,Goldberger2000mimic-physionet} for pre-training. We filter for the images with AP and PA (i.e., frontal) views. We only include the data points with the three most common pathologies (Atelectasis, Effusion, and Cardiomegaly) across the four aforementioned datasets. We also include images with No Findings to act as the negative/control class. All uncertain entries are treated as "with abnormalities". The data loading functions are modified from TorchXrayVision~\cite{cohen2020torchxrayvision1,cohen2022torchxrayvision2} and we use its downsized version of NIH and PC datasets. 
The images are central cropped and resized to $224 \times 224$ pixels. The following data augmentations are used during training: rotation ($5 \degree$), vertical and horizontal translation ($5\%$) and scaling factor interval $(0.85,\, 1.15)$. The data augmentation is performed using the RandomAffine function from TorchVision~\cite{TorchVision_maintainers_and_contributors_TorchVision_PyTorch_s_Computer_2016} (torchvision.transforms.RandomAffine).

\subsubsection*{Pipeline}
\label{ssec:pipeline}
\paragraph*{Threat Model}

In developing our collaborative training framework, we adopted an \textbf{honest-but-curious} threat model that takes the unique context and needs of our target users into account, namely hospitals and healthcare research institutes. Given that these entities have strong ethical conduct and patient-centred behaviour, coupled with their subjection to strict regulations and legal frameworks, we believe the risk of adversarial behaviour is relatively low. As such, we assume that participating hospitals will act honestly and follow the agreed-upon protocol throughout the training process. Specifically, they will compute gradients truthfully, take all necessary steps to ensure that the differential privacy guarantee is upheld (i.e., perform data points subsampling, point-wise gradient clipping, and noising as required by the protocol), as well as submit updates to the framework and perform aggregation and synchronization honestly.
That said, we acknowledge that each participant may still be curious about the input and contributions of other entities to the model. For example, an insider adversary would compute and submit gradients in the training run honestly but would attempt to infer information about other participants from the shared model updates.  There are potential privacy risks associated with such curiosity and we have implemented measures to mitigate them. For instance, our framework, \ours utilises secure aggregation, which allows leaders to aggregate the model updates collected from other entities without knowing individual updates. Instead, leaders can only view the summation of all updates, protecting the privacy of individual contributions (more in Secure Aggregation section). Furthermore, the framework would train the models to be differentially private, which limits the potential information leakage about the training data points post-training (more in the Differential Privacy section).

\paragraph*{Decentralisation}
To make the framework decentralised, we incorporate random leader selection. It is a commonly used technique to select a coordinator to facilitate some processes of a distributed system to achieve decentralisation. In \ours, specifically, at the beginning of each communication round, one participant is selected dynamically to perform the aggregation and synchronization. We assume an honest-but-curious threat model, meaning that participants will follow the protocol honestly. The leader is selected randomly, as their role is to facilitate communication among participants, rather than detect adversarial behaviour. This approach enables a decentralised framework that can be used in settings where a central server is not feasible, such as in healthcare contexts where one permanent leader (e.g., a central server) is undesirable.

\paragraph*{Secure Aggregation}
When the leader is facilitating the collaborative process of the framework, they often need to compute an aggregate value using inputs from other participants, such as the model updates. To maintain secure collaboration, it is crucial that the aggregator does not know the exact contributions of other participants while computing the aggregate computation. The \ours framework achieves this functionality by employing a well-established cryptographic protocol called Secure Aggregation (SecAgg)~\cite{Bonawitz2017SecAgg, Bell2020SecAggPlus}. 
SecAgg allows multiple participants to compute the summation of their private values without disclosing their data to others. It is commonly used in distributed settings to enable participants to collaboratively compute a function in a secure and privacy-preserving manner. Additional details about SecAgg can be found in Communication Cost of the Protocol section in Supplementary Materials. However, it is important to note that the use of SecAgg introduces communication and computation overhead to the protocol, which increases with the size of the input vector and the number of participants. Supplementary Figure 1 shows a trend of the computation and communication overhead of SecAgg. In general, the overhead increases as the increase of the size of input vectors or the number of clients. To evaluate the communication overhead for the case studies evaluated in this work, we report the total size of information transferred in SecAgg per participant and for the aggregator, as shown in Supplementary Table 1. It provides empirical evaluations of the communication cost of using SecAgg per communication round for each of the case studies. In addition, SecAgg does not incur significant computation overhead for the case studies evaluated in this work. This additional overhead should not pose a significant issue as our target clients, namely hospitals and research institutes, since they should have sufficient bandwidth to handle the additional communication.

In our protocol, we leverage SecAgg in the following three places: 
\begin{enumerate}
    \item To compute the global mean and standard deviation at the preparation stage before the training process starts, which permits each participant to normalize their private dataset without revealing it to others.
    \item To aggregate all mini-batch sizes for each iteration during the training process (Step 2 of the framework), enabling us to determine the aggregate mini-batch size for each iteration (which will be used to calculate the average gradient updates at Step 5 of the framework). 
    \item To aggregate the participants' gradient updates during the training process (Step 5 of the framework), allowing us to compute the aggregate gradient update while keeping individual clients' updates unrevealed. 
\end{enumerate}

\paragraph*{Differential Privacy}
Differential Privacy (DP) is the gold standard for reasoning about the privacy guarantee of a training algorithm. 
One of the most common working definitions of DP is $(\epsilon, \delta)$-differential privacy ($(\epsilon, \delta)$-DP): 
A randomized mechanism $M: \mathcal{D} \mapsto$ $\mathcal{R}$ satisfies $(\epsilon, \delta)$-DP if for any adjacent $D, D^{\prime} \in \mathcal{D}$ and $S \subset \mathcal{R}$
$$
\operatorname{Pr}[M(D) \in S] \leq e^\epsilon \operatorname{Pr}\left[M\left(D^{\prime}\right) \in S\right] + \delta
$$

It means that the privacy guarantee for the algorithm $M$ is bounded by $\epsilon$, but with probability $\delta$ that this guarantee may break. Hence the value for $\delta$ should be set to $0\leq \delta \leq 1$. When $\delta=0$ or negligible, it is equivalent to $\epsilon$-DP. Any $\delta \neq 0$ is a relaxation of $\epsilon$-DP. In order to achieve DP in deep learning, the de facto differential private learning algorithm is differentially private stochastic gradient descent (DP-SGD)~\cite{Abadi2016dpsgd}, as shown in~\ref{alg:dpsgd}. 

Another commonly used relaxation of DP is $(\alpha,\epsilon)$-Rényi-DP (RDP)~\cite{Mironov2017RDP}. It can be converted to $(\epsilon,\delta)$-DP for any $0<\delta<1$: 
If $M$ is an $(\alpha, \varepsilon)$-RDP mechanism, it also satisfies $(\varepsilon+\frac{\log 1/\delta}{\alpha-1},\delta)$-DP. For additive Gaussian noise (which is used in the DP-SGD algorithm), the composition rule is easier to analyse in the RDP framework hence it is commonly used to calculate the cumulative privacy budget of the DP-SGD algorithm. Therefore, for all the experiments we conducted in this paper, the privacy accounting was performed using RDP. We also follow the common practice of using a modest privacy budget (a single-digit $\epsilon$).

\begin{algorithm}
\caption{DP-SGD~\cite{Abadi2016dpsgd}}\label{alg:dpsgd}
\begin{algorithmic}[1]
\Require Dataset $D$, Mini-Batch Size $B$, Clipping Norm $C$, Noise Multiplier $\sigma$, Model Parameters $W$, Loss Function $L$, Learning Rate $\eta$

\State $W_0 \gets \texttt{RandomInit}()$
\For{$t \gets 0,\dots,T-1$} \Comment{{ \footnotesize Training Steps}}

\State {{Sample $\texttt{Mini-batch}$ } {from } $D$ with Mini-batch size $B$}

\For{{$x_b \in \texttt{Mini-batch} $}} 
\Comment{{\footnotesize Iterate over every data point in the mini-batch}}
\State {$g_b = \nabla_W L(W_{t}, \, x_b) $} \Comment{{ \footnotesize Per-example gradient calculation}}
\State {$\Bar{g_b} = g_b / \max (C^{-1} ||g_b||_2, \, 1)$} \Comment{{ \footnotesize Per-example gradient clipping}}
\EndFor
\State \textcolor{red}{$g = \frac{1}{||B||}(\sum_b \Bar{g_b} + \mathcal{N}(0, \, (C\sigma)^2))$}  \Comment{{\footnotesize Add calibrated Gaussian Noise}}
\State $W_{t+1} \gets W_{t} - \eta g$
\EndFor
\State \Return $W_T$
\end{algorithmic}
\end{algorithm}

\paragraph*{Framework (detailed)}
Setup: Suppose there are $H$ hospitals/research institutes that wish to collaborate and learn from each other's datasets. For each participant $h$, their private dataset is denoted by $$\Dataset_h = \{x_h^1, x_h^2, ..., x_h^{||\Dataset_h||}\}, \forall h \in [H]$$. 

Preparation: To initiate the training process, a random participant is selected as the leader who coordinates the initial setup. All participants would communicate the size of their private datasets $||\Dataset_h||$ to the leader and determine the sampling rate $p = \frac{B}{\sum_{h} ||\Dataset_h||}$. Here $B$ is the desired aggregate mini-batch size, which is the sum of the mini-batch sizes of all participants. The leader uses secure aggregation to compute the aggregate mean and variance from all private datasets and sends them back to each participant to normalize their private data during training. In the subsequent process, we will overload the notation $\Dataset_h$ to represent the standardized/normalized of private dataset participant $h$. 

It is worth noting that there are standard techniques available for computing the mean and variance with differential privacy guarantees to limit the privacy leakage from using these statistics. However, the privacy leakage resulting from using a global mean and variance is minimal compared to that from the sharing of gradient updates that happens at later steps of the framework. Therefore, we did not consider their privacy implications in our analysis. Finally, the leader initializes the model weight $W_0$ and distributes it to all other participants to start the training process.

\textbf{Step 1.} For communication round $t$, select one of the participants to be the leader. 

\textbf{Step 2.} For each of the participants (indexed with $h$), sample from normalized private dataset with per-point probability $p$ to get a mini-batch of data points $B_h^t = [x_h^i] \subset \Dataset_h$ of size $||B_h^t||$. The selected leader will use Secure Aggregation to  aggregate the individual mini-batch sizes and get $||B^t|| = \sum_h ||B_h^t||$. The privacy leakage for sharing the mini-batch sizes is negligible compared to the leakage from the gradient updates. So in our privacy analysis, we would ignore this. 

\textbf{Step 3.} Each of the participants follows Algorithm~\ref{alg:ind_training} to get the clipped and noised private gradient $\tilde{g}_h^t$.

\begin{algorithm}
\caption{Individual Participant Training}\label{alg:ind_training}
\begin{algorithmic}[1]
\Require Mini-batch of Dataset $B_h^t$, Communication Round $t$, the current model state $W_t$, the Clipping Norm $C$, Noise Multiplier $\sigma$, the Aggregate Mini-batch Size $||B^t||$, Loss Function $L$

\For{$x_h^i \in B_h^t$} \Comment{{\footnotesize Iterate over every data point in the mini-batch}}
\State {$g_h^t(x_h^i) = \nabla_W L(W_t, \, x_h^i) $} \Comment{{ \footnotesize Per-example gradient calculation}}
\State {$\Bar{g}_h^t(x_h^i) = g_h^t(x_h^i) / \max (\frac{||g_h^t(x_h^i)||_2}{C}, \, 1)$} \Comment{{ \footnotesize Per-example gradient clipping}}
\EndFor
\State \textcolor{red}{$\tilde{g}_h^t = \sum_i \Bar{g}_h^t(x_h^i) + \mathcal{N}(0, \, \frac{||B_h^t||}{||B^t||}(C\sigma)^2)$}  \Comment{{\footnotesize Add calibrated Gaussian Noise}}
\State \Return $\tilde{g}_h^t$
\end{algorithmic}
\end{algorithm}

\textbf{Step 4.} All the participants send their private gradients $\tilde{g}_h^t$'s to the leader. 

\textbf{Step 5.} The leader uses Secure Aggregation to aggregate the private gradient $g^t = \frac{1}{||B^t||}\sum_h \tilde{g}_h^t = \frac{1}{||B^t||}\sum_h \sum_i \Bar{g}_h^t(x_h^i) + \mathcal{N}(0, \, (C\sigma)^2)$, which is equivalent to line 7 of Algorithm~\ref{alg:dpsgd}. Then the leader performs the gradient update $W_{t+1} = W_t - \eta g^t$, where the $\eta$ is the learning rate. This is equivalent to performing standard DP-SGD on the aggregate dataset that combines all participants' private datasets. 
Note that it is crucial that the leader should only be able to see the aggregated update without access to the contribution from each participant, $\tilde{g}_h^t$. This allows the collaborative framework to reach distributed DP by adding a relatively small amount of noise (line 4 of Algorithm~\ref{alg:ind_training}) from each participant. Revealing the intermediate gradient updates would not provide the same guarantee as the aggregate noise, resulting in a lower overall privacy guarantee for the framework.   

\textbf{Step 6.} All participants synchronize and update their model state with the leader. 

\textbf{Step 7.} The new leader for the next round is selected. Step 1 to 7 is repeated until the training process finishes.

\subsubsection*{Privacy Analysis}
Given our honest-but-curious threat model, each participant in our protocol will honestly sample the data points according to the sampling rate, and the leader will honestly use secure aggregation to compute the summation of the participants' model updates. All intermediate model states revealed to the leader and then shared with other participants during training are already differentially private, making it hard for curious participants to access other participants' information. These intermediate models have the same privacy guarantee as if we are performing DP-SGD on the aggregate dataset with the same DP hyperparameters (e.g., the sampling rate, noise multiplier $\sigma$, and the number of iterations). By doing so, the models trained by \ours are able to achieve distributed DP (DDP). This is the key difference between \ours framework and PriMIA, which uses local DP. Although local DP provides privacy protection in a less constraining threat model, it adds more noise than DDP for the same privacy guarantee, resulting in a bigger performance-privacy trade-off. This often makes local DP approaches impractical to deploy. 

To ensure that the DDP guarantee holds, the participants in \ours are required to synchronize and aggregate every single iteration of training. It introduces a relatively large overhead in terms of communication. However, since we are assuming a cross-silo scenario, where the number of participating clients is small, and each client possesses a relatively large amount of data points and computing resources, every participating hospital is expected to have a bandwidth of sufficient capacity to facilitate the communication and aggregation.

\subsubsection*{Computation and algorithms}
\label{ssec:comp_and_algo}
All of the experiments are implemented in PyTorch~\cite{pytorch}. 

\paragraph*{Multilayer Perceptron (MLP)}
Multilayer Perceptron is a type of fully connected neural networks. For GEMINI study, we use an MLP with the following hyperparameters: input layer with 436 neurons, 4 hidden layers with 300, 100, 50, and 10 neurons respectively, and an output layer with 1 neuron. We use rectified linear unit (ReLU) as the activation function. To prevent overfitting, we use a weight decay of 0.0002. We used sigmoid activation after the output layer and binary cross-entropy (BCE) loss function. 
For the single-cell study, we use an MLP with the following hyperparameters: input layer with 15,558 neurons, 2 hidden layers with 1000 and 100 neurons respectively, and an output layer with 4 neurons for the 4 class labels. We use ReLU as the activation function. We use the multiclass cross-entropy loss function to perform the training. 

\paragraph*{Deep Convolutional Neural Networks}

We use DenseNet121~\cite{Huang2017densenet} model architecture for all our experiments on the pathology identification task using the chest radiology datasets. We apply transfer learning to finetune the model weights from model states pre-trained on MIMIC-CXR~\cite{Johnson2019mimic, Johnson2019mimic-physionet, Goldberger2000mimic-physionet} and ImageNet~\cite{deng2009imagenet}~\cite{ziegler2022primia_chexpert}. Conventionally, DenseNet architecture make uses of Batch Normalization (BN) layers, which keeps track of the moving average and standard deviation of the mini-batches. This layer is not allowed when training with DP-SGD since DP-SGD requires to bound the per-example gradient contribution. Hence in our experiments, we freeze the BN layers and use the pre-trained weights of those BN layers during our training.

\paragraph*{Logistic Regression}
For GEMINI study, we demonstrate the performance of our framework on logistic regression. It is implemented by using a one-layer MLP followed by Sigmoid activation function and a BCE loss function. To prevent overfitting, we also apply standard $l_2$ normalization with a weight decay of 0.0002. 

\paragraph*{Support Vector Classifier (SVC)}

For single-cell study, we also demonstrate the ability of our framework to train an SVC model. It is implemented by using a one-layer MLP followed by Multi Margin Loss. To prevent overfitting, we apply standard $l_2$ normalization with weight decay of 0.0002 during training.

\subsubsection*{Evaluation Metrics}
\label{ssec:eval_metrics}
Scikit-Learn~\cite{scikit-learn, sklearn_api} package is used to calculate the following metrics. All experiments for the three case studies are repeated with 5-fold cross validation, unless otherwise stated, where for each fold, $20\%$ of the data points from each participant are reserved as the test set to evaluate the model. 

\paragraph*{Predicting mortality of the patients}
For the patient survival/mortality prediction using GEMINI dataset, which is a binary classification task, we evaluate the Area under the receiver operating characteristic curve (AUROC), Positive predictive value (PPV), and Negative predictive value (NPV). The positive class represents the patients who die during the visit; the negative class represents the patients who survive during the visit. We use TP, FP, TN, and FN to represent true positive, false positive, true negative, and false negative respectively. The calculation of each evaluation metric is shown below: 
$$\text{PPV} = \frac{\text{TP}}{\text{TP} + \text{FP}} $$
$$\text{NPV} = \frac{\text{TN}}{\text{TN} + \text{FN}} $$

We also evaluate the F1 score for each class and compute the macro and weighted average F1 to see the effect of class imbalance. In the following calculations, $\text{TP}_c$, $\text{FN}_c$, and $\text{FP}_c$ represent the true positive, false negative, and false positive respectively for class $c \in [\text{alive, dead}]$. Let $N_c$ represent the number of cases in each of the classes. The calculation of each evaluation metric is shown below:
$$\text{F1}_c = \frac{2 \cdot \text{TP}_c}{2 \cdot \text{TP}_c + \text{FN}_c + \text{FP}_c}$$

$$\text{Macro Average F1} = \frac{\sum_c \text{F1}_c}{\sum_c 1} $$

$$\text{Weighted Average F1} = \frac{\sum_c N_c \cdot \text{F1}_c}{\sum_c N_c} $$

\paragraph*{Classifying cell types}
For the cell type classification task, we evaluate the Median F1 scores and the weighted precision and recall values. Following similar notations as above, let $\text{TP}_c$, $\text{TN}_c$, $\text{FN}_c$, and $\text{FP}_c$ represent the true positive, true negative, false negative, and false positive respectively for class $c \in [\text{alpha, beta, gamma, delta}]$. Let $N_c$ represent the number of cases in each of the classes.

$$\text{F1}_c = \frac{2 \cdot \text{TP}_c}{2 \cdot \text{TP}_c + \text{FN}_c + \text{FP}_c}$$

$$\text{Precision}_c = \frac{\text{TP}_c}{\text{TP}_c+\text{FP}_c}$$

$$\text{Recall}_c = \frac{\text{TP}_c}{\text{TP}_c+\text{FN}_c}$$

$$\text{Median F1} =  \text{Median} \{\text{F1}_c\}$$

$$\text{Weighted Precision} = \frac{\sum_c N_c \cdot \text{Precision}_c}{\sum_c N_c} $$

$$\text{Weighted Recall} = \frac{\sum_c N_c \cdot \text{Recall}_c}{\sum_c N_c} $$

\paragraph*{Identifying pathologies}
For pathology identification, which is a binary classification task, we evaluate the Area under the receiver operating characteristic curve (AUROC) for each of the three pathologies (Atelectasis, Effusion, and Cardiomegaly) as well as ``No Findings".

\subsection*{Statistical analysis}
{When presenting the model performance for the three case studies, we illustrate in the box plots the lower to upper quartile, including the median. They also include the outliers, defined as 1.5 $\times$ beyond the upper and lower quartile. The experiments are conducted with 5-fold cross-validation. Subsequently, we employ the Wilcoxon signed-rank test (one-tail) to compare the performance of different models using the five pairs of values. The test is conducted using the exact method with continuity correction, and a significance level of 0.05 is set. All values presented in the tables summarising model performance are reported as the arithmetic mean ± one standard deviation (SD). In the TPR vs. FPR plots demonstrating the model's robustness to Membership Inference Attacks, we visually represent the arithmetic mean along with a $95\%$ confidence interval derived from 5 runs. Their captions and legends accompanying these visualizations are expressed as the arithmetic mean ($\pm$ SD).}

\subsection*{Ethics statement}
GEMINI data are collected with approval from the Research Ethics Boards of all participating hospitals and this analysis was approved by Clinical Trials Ontario with the Unity Health Toronto Research Ethics Board (REB) acting as the board of record {(REB\# 20-216 and REB\# 15-087)}. {We received a waiver of informed consent from the REBs of participating institutions because of the large, retrospective nature of the data collection. Our research processes are conducted in full compliance with our approved REB protocols.} We use the scRNA-seq data of human pancreas preprocessed by previous study~\cite{Wang2022ocat}, collected from ~\cite{baron2016single_cell,muraro2016single_cell,segerstolpe2016single-cell,wang2016single-cell,xin2016rna} (Gene Expression Omnibus accession numbers GSE85241, E-MTAB-5061, GSE84133, GSE83139, and GSE81608 respectively). We use the chest X-Ray datasets from previous studies: National Institute of Health (NIH)~\cite{Wang2017nih-chestx-ray8}, PadChest (PC)~\cite{bustos_padchest_2020}, CheXpert (CheX)~\cite{irvin2019chexpert}, and MIMIC-CXR~\cite{Johnson2019mimic,Johnson2019mimic-physionet,Goldberger2000mimic-physionet}.

\subsection*{Role of the funding source}
{The funding source had no involvement in study design, data collection, data analysis, interpretation of data, writing of the manuscript, or the decision to submit the paper for publication.}

\section*{Results}
\label{sec:results}
\subsection*{\ours predicts mortality of patients admitted to hospitals using EHR }
\label{subsec:caseA}

\begin{figure}[H]
\centering\hspace{-5mm}
\includegraphics[width=\linewidth]{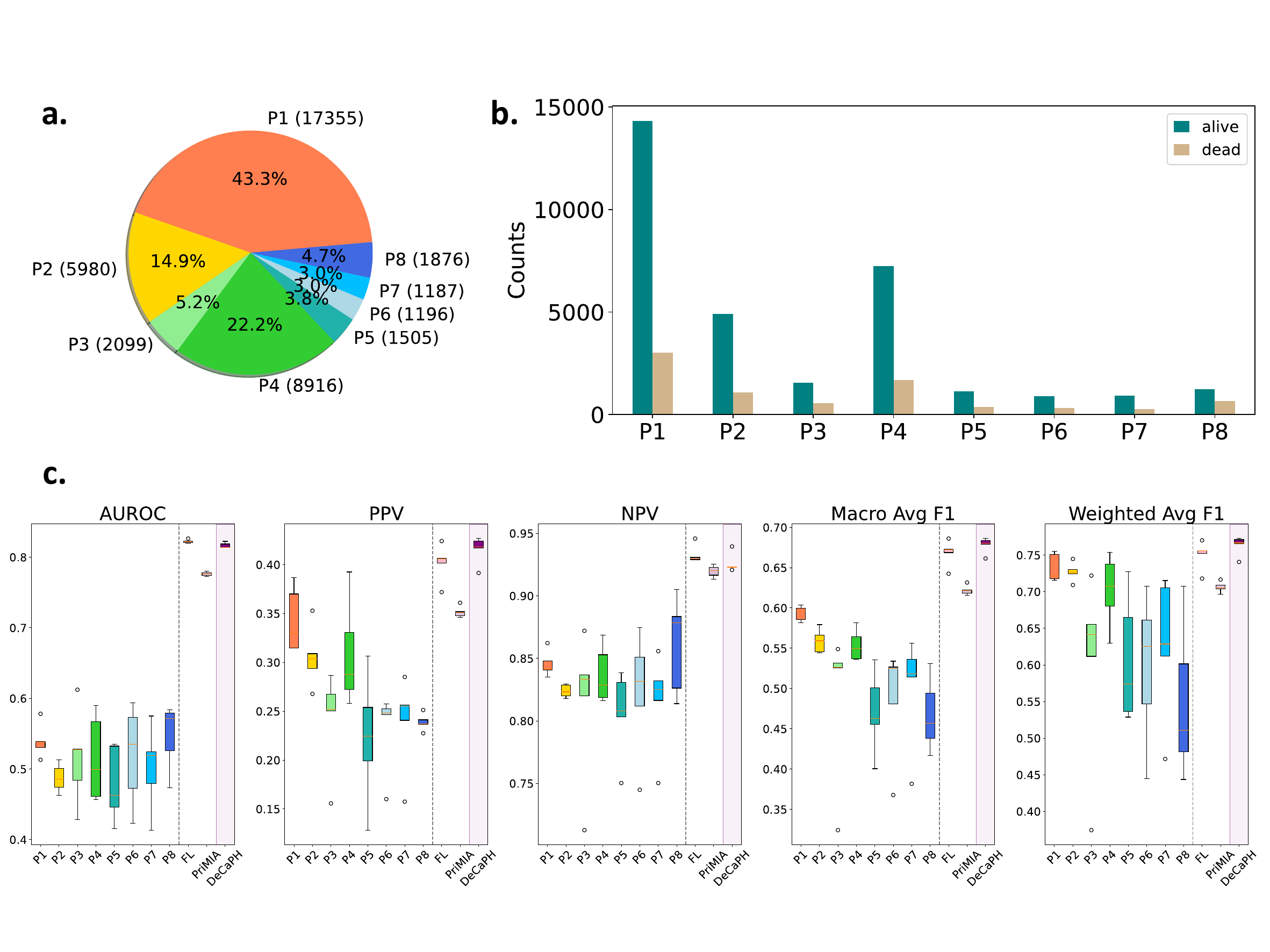}
\caption{ 
\textbf{\ours to predict mortality using EHR. }
(\textbf{a}), the number of health records available at each participating hospital {($P_1, P_2, ..., P_8$)}. (\textbf{b}), ``alive" vs. ``death" cases at each hospital. (\textbf{c}), the performance of models trained using the private datasets at each silo and models trained with all datasets using FL, PriMIA, and our \ours (highlighted in purple). The experiments are repeated with 5-fold cross-validation. The figures show the first quartile, median, and third quartile, as well as the outliers ($1.5 \times$ interquartile range below or above the lower and upper quartile.) {We perform a Wilcoxon signed-rank test (one-tail) with continuity correction using exact method to compare the performance of models trained with \ours to those trained with PriMIA for each of the evaluation metrics. The alternative hypothesis is that models trained with \ours have higher scores. The p-values are $< 0.05$ for all metrics except for NPV. }}
\label{fig:results_case_a}
\end{figure}

Our first case study analyses a dataset prepared from the GEMINI initiative~\cite{Verma2017gemini}. The dataset includes 40,114 unique hospital visits (collected from 8 hospitals) for adults admitted to a general internal medicine service from April 1, 2015 to January 23, 2020. We aim to train an ML model that can predict a patient's mortality during a hospital visit. This information has diverse uses, including clinical risk prediction and patient triaging as well as risk adjustment for research and quality measurement applications. During training, each hospital serves as one participant ($P_i$) in \ours framework and each of them has access only to its private training data points. The number of health records available at each hospital is shown in Figure~\ref{fig:results_case_a}a, and the number of mortalities is shown in Figure~\ref{fig:results_case_a}b. Since the two classes (``alive'' v.s. ``dead'') for the task are imbalanced,  the cases with the label ``dead'' are replicated three times to roughly match the number of cases with the ``alive'' label. Note that the privacy bound for training with DP depends on the data point subsampling rate, $p$; replicating the minority class in the training datasets would increase the probability of sampling data points from this minority class. Even though this practice in principle weakens the bound on privacy leakage provided by the differential privacy analysis, we will show in the Results section that models trained with \ours are still less vulnerable than those trained with FL because \ours framework provides a privacy guarantee whereas models trained with FL cannot.  

Recall that the primary goal of \ours is to enable multiple parties to collaborate and train a model to have better performance than those trained using only one of the private datasets available at each hospital; meanwhile, the framework needs to make sure the collaboratively trained models conform to DP. To evaluate the effectiveness of \ours, we first systematically compare the performance of \ours-trained models with models trained using one of the private datasets or previous collaborative training frameworks. Later in the Results section, we empirically evaluate the robustness of \ours-trained models against privacy attacks. 
For the first case study, we compare the performance of the model trained with only one participant’s private dataset, and the models trained with all eight private datasets using FL~\cite{mcmahan2017fl}, PriMIA~\cite{Kaissis2021primia}, and \ours. We use a multi-layer perceptron (MLP) as the model architecture~\cite{Vaid2020fl_ehr} and stochastic gradient descent (SGD) optimizer for the training, as the results are presented in Figure~\ref{fig:results_case_a}c. We also repeat the same experiments using a one-layer linear model to run logistic regression for this task~\cite{Vaid2020fl_ehr}. The results are presented in Supplementary Figure 2 and Supplementary Table 5, with similar qualitative results as for MLP models. 
The models are evaluated for a few different metrics, Area Under Receiver Operating Characteristic curve (AUROC), the true positive value (PPV), the true negative value (NPV), the Macro Average F1, and the Weighted Average F1. The threshold is determined using Youden's J Statistic for each fold.  It is observed that the models trained with FL and \ours framework consistently perform better than models trained with only the private dataset at that silo. The models trained with \ours are privacy-preserving (with a privacy budget of $\epsilon=2.0$), whereas the models trained with FL do not provide any privacy guarantee. In addition, by carefully calibrating the privacy-related hyperparameters, the performance of models trained with \ours is on par with those trained with FL. The average performance degradation of the models trained with \ours compared with that of FL is less than 1\% in all metrics, as shown in Supplementary Table 4. With a small loss of utility, the models trained with \ours are significantly more robust to privacy attacks, as later evaluated in the Results section.

We also observe that the test performance of the models trained by PriMIA is lower than those trained with \ours when using the same privacy budget ($\epsilon=2.0$). PriMIA is a differentially private implementation of FL. Each client would use DP-SGD to train their local models so that their updates submitted to the central server would already be differentially private, which means each client would perform the computations without considering the potential contributions from other participants. Hence, some clients may reach the target privacy budgets in less iterations than others and terminate training. Usually, only one party remains in training towards the end of the training phases. But training the model with only one participant’s data would cause the model parameter to forget about the knowledge of other clients (just like the catastrophic forgetting in transfer learning). Also, since PriMIA runs local DP-SGD without considering the potential contributions from other participants, it tends to add more noise than needed for the particular privacy budget, hence it might also cause performance degradation. More details about PriMIA and other existing frameworks are provided in the Existing frameworks section of the Supplementary Materials.

\subsection*{\ours classifies cell types in single-cell human pancreas studies }
\label{subsec:caseB}

\begin{figure}[H]
\centering\hspace{-5mm}
\includegraphics[width=\linewidth]{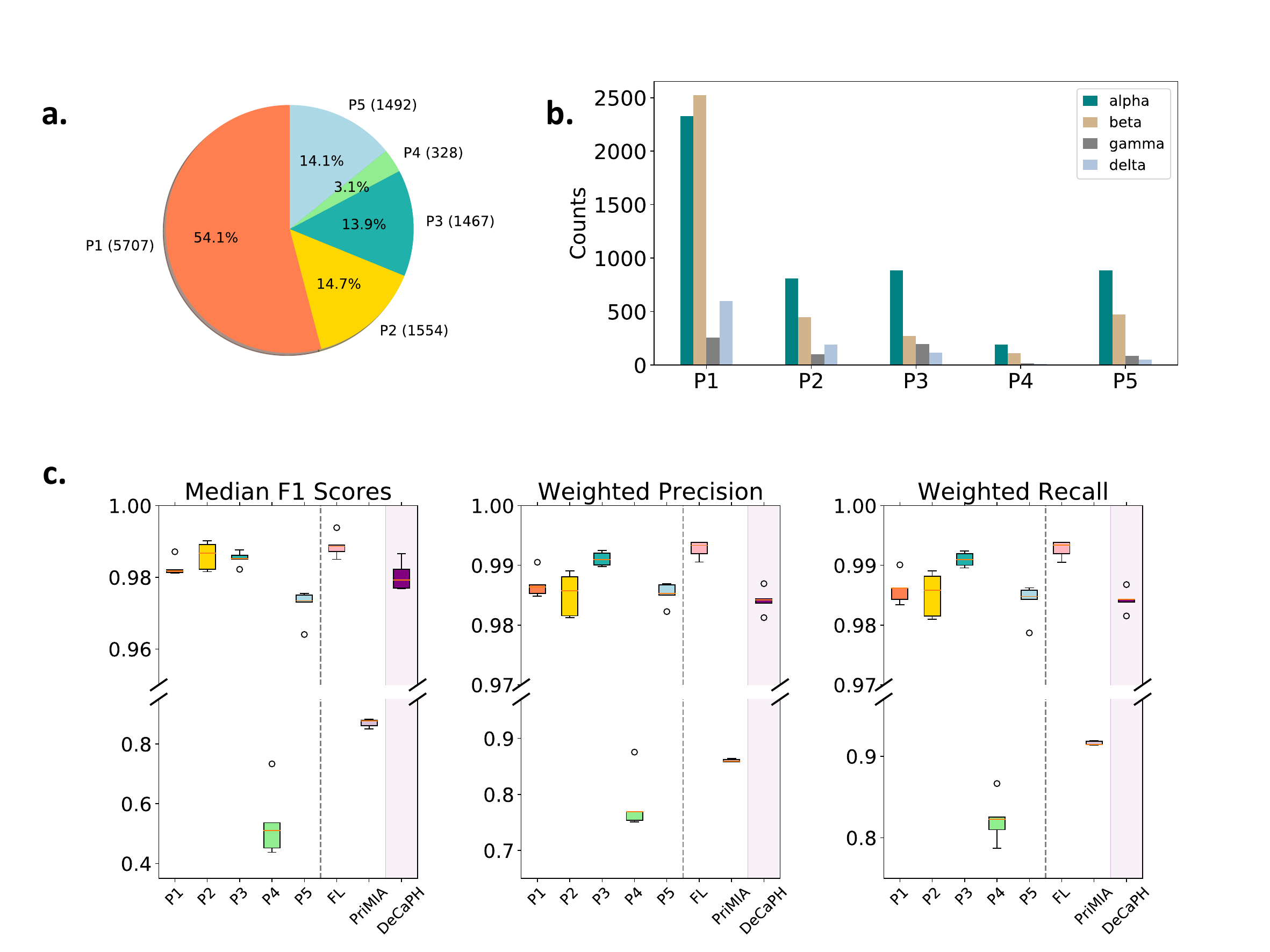}
\caption{ 
\textbf{\ours to classify cell types using single-cell human pancreas dataset. }
(\textbf{a}), the number of data points available in each participating study, {($P_1, P_2, ..., P_5$)}. (\textbf{b}), the proportion of the classes in the datasets. (\textbf{c}), the performance (with 5-fold cross-validation) of the models trained using the private dataset of each study and the models trained with all datasets using FL, PriMIA, and \ours (highlighted in purple). We break the axis for better visualization. The figures show the first quartile, median, and third quartile, as well as the outliers ($1.5 \times$ interquartile range below or above the lower and upper quartile.) {We perform a Wilcoxon signed-rank test (one-tail) with continuity correction using exact method on performance of models trained with \ours and PriMIA for each of the evaluation metrics. The alternative hypothesis is that models trained with \ours have higher scores for that metric. The p-values are $< 0.05$ for all metrics. }}
\label{fig:results_case_b}
\end{figure}

In the second case study, the goal is to classify different cell types by training models using datasets collected from five distinct studies. Each study is treated as a separate participant, denoted as $P_i, \, i\in [1,5]$. The sizes of the private datasets from each study are shown in Figure~\ref{fig:results_case_b}a. In this study, we only consider the 4 common cell types across all studies for this classification task, namely alpha, beta, gamma, and delta. The distribution of the cell types for each individual dataset is visualized in Figure~\ref{fig:results_case_b}b. Similar to the first case study, we compare the test performance of models trained with only the private dataset available at each silo and the models trained with FL, PriMIA, and \ours framework with 5-fold cross-validation. Following the model architecture used in previous single-cell analysis~\cite{Ma2021single_cell_mlp}, we employ an MLP model for this case study. The performance of the models is evaluated using three evaluation metrics: median F1 scores, average precision, and average recall, as shown in Figure~\ref{fig:results_case_b}c. 

The results show that the models trained with only one of the private datasets can also reach close-to-perfect test performance, except for $P_4$, which has little data available, resulting in significantly worse performance than models trained with the private dataset at other silos. \ours and FL significantly outperform the models trained with private data from $P_4$ and perform similarly to the models trained with only the private data from other studies. 
In addition, it is observed that when using PriMIA, if the mini-batch sampling rates at different participants are not the same, for example, all the participants use the same mini-batch size locally but possess varying dataset sizes, some participants would use up their budget in fewer iterations than others. This effect is more dominant when one of the participants has significantly more data points (in this case, $P_1$) than other participants. $P_1$ would be able to train for more iterations than other participants causing the final model to learn less about other datasets and bias towards the data distribution of $P_1$. 
{The overall qualitative comparison between FL, PriMIA, and \ours is similar to the previous case study.} 
More details about the performance of the MLP models are summarised in Supplementary Table 6.

Note that the privacy budgets for different tasks are chosen specifically for different datasets to ensure the privacy-preserving model has a good performance. A modest privacy budget (single-digit $\epsilon$) is used following the consensus in the literature~\cite{Abadi2016dpsgd}. In this case study, a privacy budget of $\epsilon=5.65$ is used.

We conduct additional experiments on the same task but through training a support vector classifier (SVC)~\cite{Alquicira-Hernandez2019scPred}. The results are included in Supplementary Figure 3 and Supplementary Table 7. The trend is similar to the case where MLP models are used. Moreover, there are no noticeable batching effects when training these models for the classification task, which is consistent with previous studies on inter-dataset performance of such models~\cite{Abdelaal2019comparisonsinglecell}.

\subsection*{\ours identifies pathologies from human chest radiology images}
\label{subsec:caseC}

\begin{figure}[H]
\centering
\includegraphics[width=\linewidth]{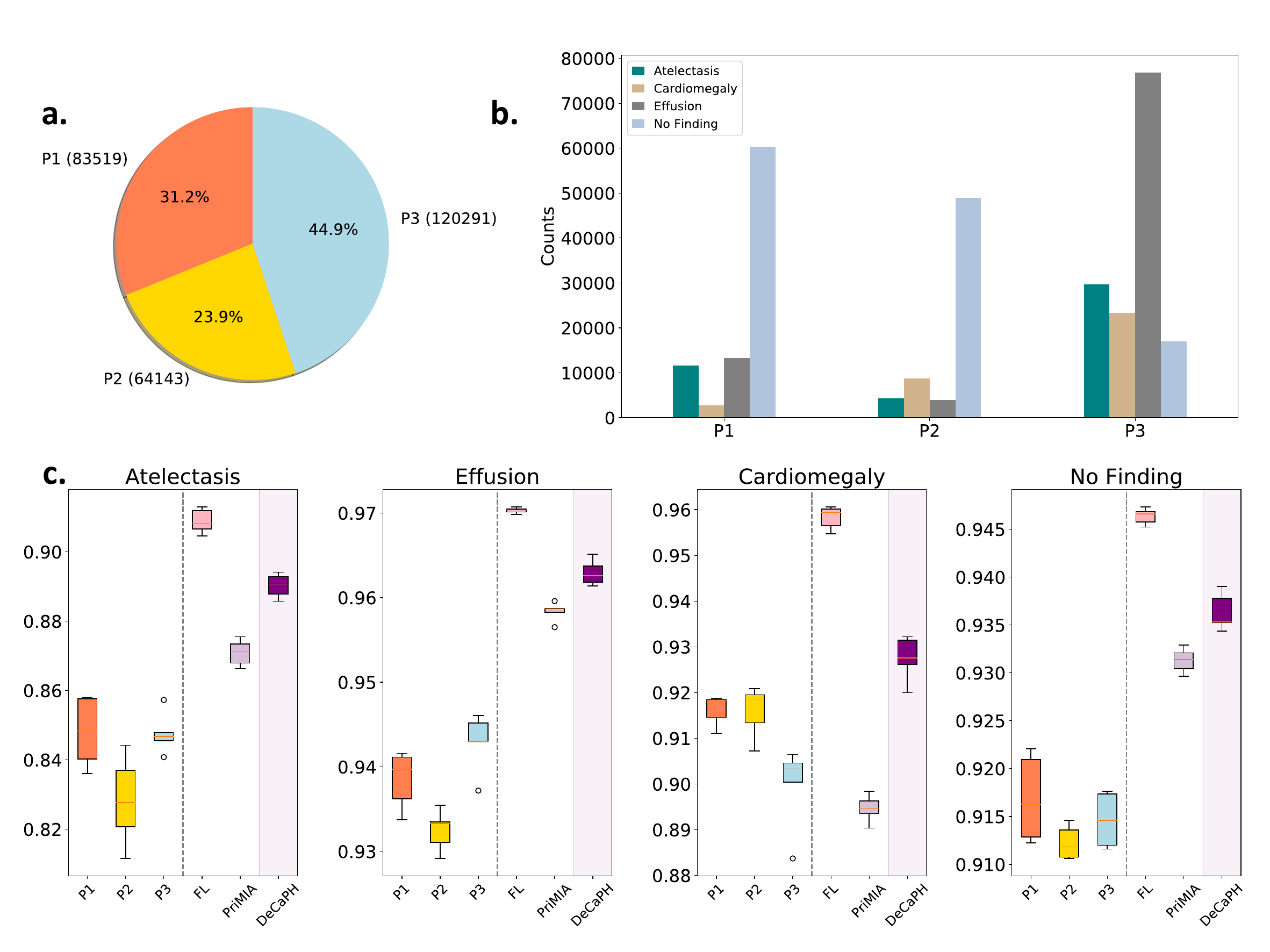}
\caption{
\textbf{\ours to identify pathologies from human chest radiology images}
(\textbf{a}), the sizes of the datasets available in each study, {($P_1, P_2, P_3$)}. (\textbf{b}), the class distribution of the datasets. (\textbf{c}), the performance on AUROC for the four output labels (with 5-fold cross-validation) of the models trained using
the private dataset of each study and the models trained with all datasets using FL, PriMIA, and
\ours (highlighted in purple).  The figures show the first quartile, median, and third quartile, as well as the outliers ($1.5 \times$ interquartile range below or above the lower and upper quartile.) {We perform a Wilcoxon signed-rank test (one-tail) with continuity correction using exact method on performance of models trained with \ours and PriMIA for each of the pathologies and ``No Finding". The alternative hypothesis is that models trained with \ours have higher AUROC scores. The p-values are $< 0.05$ for all three pathologies and ``No Finding". }}
\label{fig:results_case_c}
\end{figure}

For the third study, we demonstrate the versatility of the \ours framework by applying it to multi-label classification tasks on three human chest radiology datasets. Unlike the previous two tasks, which use tabular datasets, this analysis uses X-ray imaging datasets. Additionally, while the previous experiments only involved binary or multiclass classification tasks, this analysis performs multilabel classification, where each input can have multiple output classes (e.g., multiple pathologies can be identified from one X-ray image). The model is trained to predict whether the patient has the following three pathologies: Atelectasis, Effusion, and Cardiomegaly, or no abnormality is noted (i.e., the image is labelled as "no finding"). The size of the filtered datasets and the distribution of the classes are presented in Figure~\ref{fig:results_case_c}a and Figure~\ref{fig:results_case_c}b, respectively. The architecture used for this study is a deep convolution neural network (CNN), which is a commonly used architecture for image recognition tasks such as these~\cite{Almezhghwi2021chest_radiology_cnn}. Specifically, we used a DenseNet121 architecture~\cite{Huang2017densenet}. The model has four outputs, where each output is a binary classification indicating the presence or absence of the pathology in the image. In other words, the four outputs of the model predict if the X-ray image has Atelectasis, Effusion, Cardiomegaly, or No Finding. 

To improve the model's performance and training efficiency, we employ transfer learning, where the model's state is initialized with weights pre-trained on the MIMIC-CXR dataset with the same four outputs. Transfer learning~\cite{pan2010transfer_learning} and pre-trained models are widely-used strategies in computer vision, as the pre-trained weights contain low-level features of the images; these low-level features can be transferred to improve the model's performance on new datasets, especially when the dataset for the downstream task is relatively small or private, and needs to be trained with differential privacy~\cite{pan2010transfer_learning,de2022unlocking}\footnote{\cite{de2022unlocking} is a pre-print.}. 

For the evaluation, we compare the performance of the models trained with only the private dataset available at each silo to those trained using all datasets with FL, PriMIA, and \ours frameworks. We evaluate the AUROC scores for the four output labels, as shown in Figure~\ref{fig:results_case_c}c. We set $\epsilon=0.62$ for both PriMIA and \ours. The results demonstrate that models trained on all datasets (i.e., models trained with FL, PriMIA, and \ours) outperform those trained on individual private datasets available at each silo. Furthermore, the models trained with \ours show less utility degradation than those trained with PriMIA when using the same privacy budget. Overall, the models trained with \ours guarantee privacy with little utility loss (no more than $3.2\%$), as shown in Supplementary Table 8.

We also evaluate the scenario where the initial model is pre-trained with ImageNet~\cite{gundel2019transfer_on_xray,deng2009imagenet} present the results in Supplementary Figure 4 and Supplementary Table 9. However, we observed a larger utility degradation for models trained with privacy guarantee (i.e., models trained with PriMIA and \ours) compared to the scenario where the models are pre-trained with MIMIC-CXR. This result suggests that it is harder for models trained with DP to converge when the pre-trained model is trained on a dissimilar dataset like ImageNet, compared to a more similar dataset like MIMIC-CXR. This is because DP training involves constant gradient clipping and noise addition, making it harder for the model to converge from an initial state trained on a dissimilar task. We observe a consistent trend where models trained with PriMIA experienced larger utility degradation than those trained with \ours, as some participants may terminate training due to using up the privacy budget or adding more noise than necessary.

\subsection*{Models trained with \ours are more robust to privacy attacks}
\label{subsec:robustness}

\begin{figure}[H]
\centering
\includegraphics[width=\linewidth]{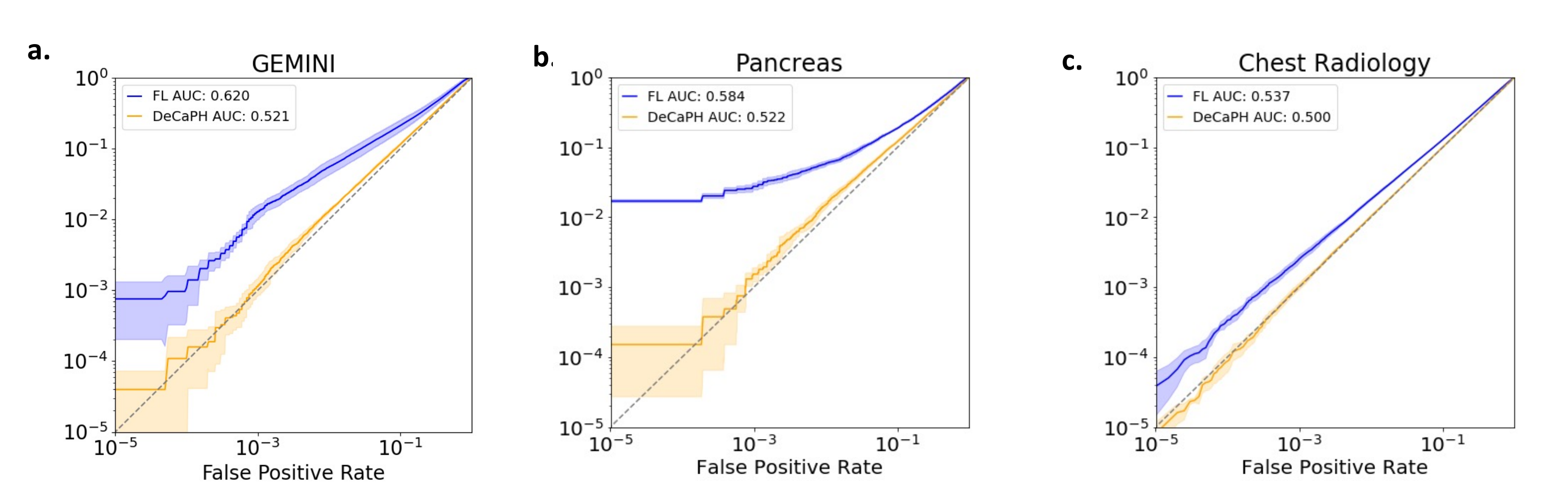}
\caption{\textbf{Models trained with \ours are more robust to Membership Inference Attacks.} We perform Membership Inference Attack on models trained with \ours vs. FL for the three case studies. The models trained with \ours (Ours) are differentially private. The models trained with FL are not privacy-preserving. The target models are trained five times to plot the {$95\%$} confidence interval. (\textbf{a}), for GEMINI, the AUROC for FL is  $0.620\pm0.043$ and that for \ours is $0.521 \pm 0.003$. (\textbf{b}), for single-cell human pancreas, the AUROC for FL is  $0.584\pm0.009$ and that for \ours is $0.522 \pm 0.004$. (\textbf{c}), for chest radiology, the AUROC for FL is $0.537\pm0.001$ and that for \ours is $0.500 \pm 0.001$; {mean $\pm$ SD.}}
\label{fig:mia}
\end{figure}

Thus far, we have compared the utility of the models trained with \ours and reported the DP budget. {In this section, in addition to the theoretically privacy analysis presented earlier, we perform an ablation study to empirically demonstrate the value of integrating a privacy-preserving mechanism in collaborative framework.} To assess the effectiveness of the privacy-preserving mechanism, we conduct a membership inference attack (MIA)~\cite{shokri2017membership,carlini2022membership}, the standard method to evaluate how much private information practically leaks from the model. The adversary’s goal is to predict if a given data point is a member of the training dataset used to train the target model. Predicting membership can leak private information in at least two ways. First, membership in the dataset can be sensitive if, for example, the dataset contains records of patients that have a specific condition: being part of the dataset implies the individual has this medical condition. Second, membership inference is often used as a primitive to mount other attacks such as training data reconstruction attacks. The success of MIA offers a way to analyse the privacy guarantees provided by the training algorithms in a way that is complementary to a differential privacy analysis. In \ours, the adversary could be a curious participant who may actively find out information about other participants during the training phase. Or an adversary could be anyone (e.g., the participants, the general public, etc.) who has access to the final model state after deployment of the model. 

For each of the three case studies, we evaluate the vulnerability of two target models, one is the final model trained with \ours whereas the second is that trained using FL without any privacy guarantees. To ensure a fair comparison, the FL target models here are trained using the same mini-batch sampling rates and the synchronization frequency as \ours would use, i.e., the only difference is the absence of gradient clipping and noising in FL while computing steps of gradient descent. We use the state-of-the-art MIA technique, Likelihood Ratio Attack (LiRA)~\cite{carlini2022membership}, to predict the membership information of the two target models. 
More details about LiRA are provided in the Membership Inference Attack section in Supplementary materials. To evaluate the success of the attack on the target models, we follow the recommendations of~\cite{carlini2022membership} to plot the ROC curve, which shows the True Positive Rate (TPR) versus the False Positive Rate (FPR) of the adversary's prediction, and focus on the low-FPR regime. Also, we report the AUROC for both target models. For consistency, we use the same model architectures and training setups as previous experiments for LiRA. The comparison of the model vulnerabilities for the three case studies is shown in Figure~\ref{fig:mia}. It is observed that the models trained with \ours are much less vulnerable to the attack compared to models trained with FL. 

We present additional results on different model architectures in Supplementary Figures 5 and 6. For instance, when we use a one-layer linear model to run logistic regression on the GEMINI dataset, we observe that the attack success rate is similar for both models trained with FL and \ours, as shown in Supplementary Figure 5. In addition, the attack is much less successful on the linear model compared to the MLP models, especially when trained with FL (without any privacy guarantee). 
This may suggest that the limited capacity of the one-layer model may make it less prone to overfitting, resulting in better privacy protection. This finding is also reflected by the slightly lower model utility of the linear models compared to the MLP models (see comparison in Figure~\ref{fig:results_case_a}c and Supplementary Figure 2). This also suggests that when model utility is comparable, using a simpler model architecture with fewer parameters may reduce the risk of privacy leakage.

In contrast, for the pancreas dataset, we do not observe better privacy protection when using simpler model architectures, such as an SVC (shown in Supplementary Figure 6), compared to using larger MLP models, especially at the low-FPR regime. However, the overall trend is the same as with MLP models: the target models trained with FL (without any privacy guarantee) are much more vulnerable to membership inference attacks than the model trained with our \ours framework. This may be because the pancreas dataset is relatively simple and an SVC is already sufficient for the task.

\section*{Conclusion and Discussion}
\label{sec:discussion} 
We demonstrate the capability of \ours by training models on three tasks: prediction of patients mortality using EHR, classification of cell types using single-cell RNA datasets, and identification of pathologies using human chest radiology. The models trained with \ours achieve better performance than models only training on one of the private datasets available at each silo. {This demonstrates that \ours is capable of handling large variety of different data types and tasks, namely low-dimensional tabular EHR datasets, high-dimensional genomics datasets as well as imaging datasets. In addition, we used real-world cross-silo datasets it demonstrate that \ours has the potential to handle the complexity and the heterogeneity of real-world datasets, which demonstrates its potential to be deployed in real-world and in turns aid human experts.} Furthermore, we show that the models trained with \ours are more robust to privacy attacks, like membership inference attack, {to empirically demonstrate the value of adding privacy-preserving techniques in terms of protecting patients' information.}
{Overall, }\ours framework enables researchers to conduct larger-scale ML studies and train more accurate models by leveraging heterogeneous sources of data points without compromising patient privacy. Overall, our framework provides a promising solution to enable secure and private collaboration for ML research on healthcare-related topics. 

We expect future work will further strengthen \ours framework in multiple ways. First, \ours only supports horizontal integration of datasets currently, which means that the different private datasets need to have the same set of inputs and outputs. Vertically integrating datasets would allow \ours framework to extend to datasets with varying inputs and outputs. {This poses a non-trivial challenges~\cite{Xu2022private_vertical_fl}; it may require additional techniques to approximately align the data points at each silo by some common universal identifiers, e.g., patient ID, that may or may not be available. Also, such process has to prioritize confidentiality and privacy of the datasets. In addition, a more sophisticated design of the training and aggregation process is required to merge the gradients or model updates computed at each hospital are computed using different input features.
} Second, we demonstrated the feasibility of \ours framework on supervised learning, leaving room for its adaptation to unsupervised, semi-supervised, {and self-supervised learning} scenarios, {e.g., large language models for tasks like clinical note transcription. They usually involve much larger models, inherently facing larger privacy-utility trade-offs and communication overhead. Therefore, it is crucial to explore and employ more techniques to achieve good privacy-utility-communication trade-offs. Moreover, ensuring the samples generated by the models are not leaking private information of the data itself could be a research field.} {Furthermore, during deployment of the framework, more things need to be considered. E.g., ensuring secure communication among hospitals and storage of the database~\cite{Almulihi2022healcare_data_breach}, maintaining software reliability~\cite{sahu2021reliability}, safe onboarding of participants, maintaining logs of the transaction/training process, etc. }

\section*{Contributors}
{CF, AD, NP, and BW conceptualised the study and developed the methodology. CF conducted the experiments and performed the analysis. CF and LO made the visualization. LZ, AV, and FR curated the datasets used for the study. NP and BW acquired funding and supervised the study. CF wrote the original draft. CF, LZ, AV, NP, and BW reviewed and edited the manuscript. All authors read and approved the final version of the manuscript. CF and BW had verified the underlying data. }

\section*{Data sharing}
\subsection*{Code Availability}
The code is available at \url{https://github.com/cleverhans-lab/DeCaPH}.

\subsection*{Data Availability}
\subsubsection*{GEMINI}
GEMINI~\cite{Verma2017gemini} is an electronic health record dataset collected from hospitals across Ontario. In this study, we look at the patient records from eight hospitals: Humber River Hospital (HRH), St. Michael's Hospital (SMH), Markham Stouffville Hospital (MKSH), Sunnybrook Health Sciences Centre (SBK), Mount Sinai Hospital (MSH), Toronto General Hospital (UHNTG), Toronto Western Hospital (UHNTW), and St. Joseph's Health Centre (SJHC). Data cannot be made publicly available due to limitations in research ethics approvals and data sharing agreements, but access can be obtained upon reasonable request and in line with local ethics and privacy protocols, via \url{https://www.geminimedicine.ca/}. More information about data can be found in the Data Collection section of the Supplementary Materials.

\subsubsection*{Single Cell Human Pancreas} 
We use the scRNA-seq data of human pancreas collected from five different studies: Baron~\cite{baron2016single_cell}, Muraro~\cite{muraro2016single_cell}, Segerstolpe~\cite{segerstolpe2016single-cell}, Wang~\cite{wang2016single-cell}, and Xin~\cite{xin2016rna} (Gene Expression Omnibus accession numbers GSE85241, E-MTAB-5061, GSE84133, GSE83139, and GSE81608 respectively). The preprocessed version is {openly available}, provided by~\cite{Wang2022ocat}(\url{https://data.wanglab.ml/OCAT/Pancreas.zip}).

\subsubsection*{Chest Radiology}
We use the chest X-Ray datasets from National Institute of Health (NIH)~\cite{Wang2017nih-chestx-ray8} (\url{https://www.nih.gov/news-events/news-releases/nih-clinical-center-provides-one-largest-publicly-available-chest-x-ray-datasets-scientific-community}), PadChest (PC)~\cite{bustos_padchest_2020}(\url{https://bimcv.cipf.es/bimcv-projects/padchest/}), CheXpert (CheX)~\cite{irvin2019chexpert}(\url{https://stanfordmlgroup.github.io/competitions/chexpert/}), and MIMIC-CXR~\cite{Johnson2019mimic,Johnson2019mimic-physionet,Goldberger2000mimic-physionet}(\url{https://physionet.org/content/mimic-cxr-jpg/2.0.0/}). {Access to the datasets can be obtained via the links above.} For NIH and PadChest, we used the downsized version provided by TorchXrayVision~\cite{cohen2020torchxrayvision1,cohen2022torchxrayvision2}(\url{https://academictorrents.com/details/e615d3aebce373f1dc8bd9d11064da55bdadede0}, \url{https://academictorrents.com/details/96ebb4f92b85929eadfb16761f310a6d04105797}).

\section*{Declaration of interests}
The authors declare no conflict of interest.

\section*{Acknowledgements}
This work is funded by the Natural Sciences and Engineering Research Council of Canada (NSERC, RGPIN-2020-06189 and DGECR-2020-00294), Canadian Institute for Advanced Research (CIFAR) AI Catalyst Grants,  CIFAR AI Chair programs, Temerty Professor of AI Research and Education in Medicine, University of Toronto, Amazon, Apple, DARPA through the GARD project, Intel, Meta, the Ontario Early Researcher Award, and the Sloan Foundation. Resources used in preparing this research were provided, in part, by the Province of Ontario, the Government of Canada through CIFAR, and companies sponsoring the Vector Institute.

\section*{Declaration of generative AI and AI-assisted technologies in the writing process}
During the preparation of this work, the author(s) used ChatGPT in order to improve grammar and wording. After using this tool/service, the author(s) reviewed and edited the content as needed and take(s) full responsibility for the content of the publication.

\bibliographystyle{vancouver}
\bibliography{my_bib}

\section*{Figure Legends}
Fig.1: \textbf{An overview of \ours learning framework.} (\textbf{a}), flowchart of the steps for one iteration of training with \ours. At each communication round, \encircle{1} a leader is first selected to perform the \underline{aggregation of the participants' model weights}; \encircle{2} each hospital locally randomly samples a mini-batch of data points and computes their point-wise gradients; \encircle{3} each hospital locally clips the point-wise gradient vectors and adds a calibrated Gaussian Noise; \encircle{4} all participating hospitals send their local gradients to the leader; \encircle{5} the leader aggregates the gradients from all hospitals using SecAgg and outputs \underline{an updated model} that is \underline{differentially private}; \encircle{6} all participating hospitals synchronize their model state with the leader. \underline{Reiterate these steps until convergence.}  (\textbf{b}), visualization of one training iteration of \ours with three participating hospitals.

Fig.2: \textbf{\ours to predict mortality using EHR. }
(\textbf{a}), the number of health records available at each participating hospital ($P_1, P_2, ..., P_8$). (\textbf{b}), ``alive" vs. ``death" cases at each hospital. (\textbf{c}), the performance of models trained using the private datasets at each silo and models trained with all datasets using FL, PriMIA, and our \ours (highlighted in purple). The experiments are repeated with 5-fold cross-validation. The figures show the first quartile, median, and third quartile, as well as the outliers ($1.5 \times$ interquartile range below or above the lower and upper quartile.) We perform a Wilcoxon signed-rank test (one-tail) with continuity correction using exact method to compare the performance of models trained with \ours to those trained with PriMIA for each of the evaluation metrics. The alternative hypothesis is that models trained with \ours have higher scores. The p-values are $< 0.05$ for all metrics except for NPV.

{Fig.3: \textbf{\ours to classify cell types using single-cell human pancreas dataset. }
(\textbf{a}), the number of data points available in each participating study, ($P_1, P_2, ..., P_5$). (\textbf{b}), the proportion of the classes in the datasets. (\textbf{c}), the performance (with 5-fold cross-validation) of the models trained using the private dataset of each study and the models trained with all datasets using FL, PriMIA, and \ours (highlighted in purple). We break the axis for better visualization. The figures show the first quartile, median, and third quartile, as well as the outliers ($1.5 \times$ interquartile range below or above the lower and upper quartile.) We perform a Wilcoxon signed-rank test (one-tail) with continuity correction using exact method on performance of models trained with \ours and PriMIA for each of the evaluation metrics. The alternative hypothesis is that models trained with \ours have higher scores for that metric. The p-values are $< 0.05$ for all metrics. }

{Fig.4: \textbf{\ours to identify pathologies from human chest radiology images}
(\textbf{a}), the sizes of the datasets available in each study, ($P_1, P_2, P_3$). (\textbf{b}), the class distribution of the datasets. (\textbf{c}), the performance on AUROC for the four output labels (with 5-fold cross-validation) of the models trained using
the private dataset of each study and the models trained with all datasets using FL, PriMIA, and
\ours (highlighted in purple).  The figures show the first quartile, median, and third quartile, as well as the outliers ($1.5 \times$ interquartile range below or above the lower and upper quartile.) We perform a Wilcoxon signed-rank test (one-tail) with continuity correction using exact method on performance of models trained with \ours and PriMIA for each of the pathologies and ``No Finding". The alternative hypothesis is that models trained with \ours have higher AUROC scores. The p-values are $< 0.05$ for all three pathologies and ``No Finding". }

{Fig.5: \textbf{Models trained with \ours are more robust to Membership Inference Attacks.} We perform Membership Inference Attack on models trained with \ours vs. FL for the three case studies. The models trained with \ours (Ours) are differentially private. The models trained with FL are not privacy-preserving. The target models are trained five times to plot the $95\%$ confidence interval. (\textbf{a}), for GEMINI, the AUROC for FL is  $0.620\pm0.043$ and that for \ours is $0.521 \pm 0.003$. (\textbf{b}), for single-cell human pancreas, the AUROC for FL is  $0.584\pm0.009$ and that for \ours is $0.522 \pm 0.004$. (\textbf{c}), for chest radiology, the AUROC for FL is $0.537\pm0.001$ and that for \ours is $0.500 \pm 0.001$; {mean $\pm$ SD.}}

\end{document}


\maketitle

\section*{Existing Frameworks}
\label{supp_sec:existing_works}
Existing collaborative ML training frameworks, while effective in some contexts, have limitations that make them unsuitable for use in healthcare settings. For example, Federated Learning (FL)~\cite{mcmahan2017fl} is an established framework that protects the confidentiality of clients' datasets. For each communication round during the training phase of FL, (selected) clients would perform training with their local private datasets for an agreed-upon number of iterations. Then the clients would submit their model updates to a central server which would then perform the model weights aggregation—following that the server would distribute the aggregate model state to the clients and the new round starts. However, revealing the exact values of the clients' updates to the server may leak information about clients' private data. As a solution, the server could employ secure aggregation (SecAgg) so that it would compute the sum of clients' updates without knowing the values of their contributions. This would make it harder for the server to gain information about the clients' private data. However, it still does not prevent privacy leakage from the aggregated model weights. A differentially private version of FL, such as DP-FL, has been developed to protect client-level privacy~\cite{McMahan2017dpfl-rnn, geyer2017dpfl-client-level}\footnote{\cite{geyer2017dpfl-client-level} is a pre-print.}. 
These frameworks are suitable for scenarios where each client corresponds to a personal health device or a cellphone, which the client's data pertains to one individual or a family. 
However, this granularity of privacy protection is insufficient for the healthcare settings considered in this paper where hospitals and research institutes would like to collaborate. In this scenario, each hospital/research institute typically contains data points from multiple patients, and privacy must be protected at the patient-level, rather than the hospital-level. Therefore, \ours framework is designed to achieve privacy protection at the granularity of every single patient in the dataset.

PriMIA~\cite{Kaissis2021primia} is a collaborative ML training framework that combines FL with SecAgg to aggregate the local updates on the server side and utilises Differentially Private Stochastic Gradient Descent (DP-SGD)~\cite{Abadi2016dpsgd} in the local client training process. This framework addresses the issue of lacking patient-level privacy protection in FL, but still requires a central server. The existence of a central server may not be suitable for healthcare settings since the server needs to perform the extra computation to perform the aggregation and coordinate the synchronization of all the clients. The framework would be more scalable if this leadership role can be rotated across all participants so that the extra computation overhead is also distributed. Additionally, the privacy budget for DP-SGD training depends on the data point subsampling rate, which can lead to faster budget consumption for higher subsampling rates. As different parties typically have varying numbers of data points, if the mini-batch sizes and subsampling rates for different entities are not set carefully, some parties may use up their budget sooner than others. This could result in only one party training towards the end of the training phases, causing the aggregate model to forget about other parties' contributions. Furthermore, since PriMIA adds noise locally without considering the contribution from all other participating entities, it may add more noise than necessary, creating a trade-off between privacy budget and performance degradation that could be too significant. Therefore, when designing \ours framework, the goal is to add the minimal amount of noise necessary to ensure the privacy guarantee is met while minimizing the utility degradation of the models. 

Swarm Learning (SL)~\cite{Warnat-Herresthal2021swarmlearning} is a decentralised version of federated learning that replaces the central server with blockchain technology to coordinate the synchronization and aggregation of model updates. This decentralisation gives the framework more transparency and helps prevent a single-point failure in the server that will impact the entire training process. However, SL, like FL, does not provide any provable privacy protection. In other words, there is no theoretical bound on the potential privacy leakage from the final model or the sharing of the model updates during the training process. As a result, SL may not be suitable for some healthcare applications where the training data contains highly sensitive information (e.g., patient's health record) and require to be protected protection with a theoretical upper bound on the privacy leakage. 

There is another class of collaborative learning frameworks where participants share model predictions instead of model updates. An example framework is the Private Aggregation of Teacher Ensembles (PATE)~\cite{papernot2018pate}. The PATE framework would require each of the participants to train a teacher model using their individual private datasets. At inference time, the teacher models are queried to provide their predictions of an input. The teachers' predictions are aggregated and their votes are applied with noise to protect the privacy of the participants' datasets. It is a useful framework when each of the participants has enough data to train their own local model to a reasonable utility but still wants to collaborate to enhance the overall performance of their models on new queries. An improved version of PATE, Confidential and Private Collaborative (CaPC) learning, applies secure multiparty computation to aggregate the teacher models' predictions and add  noise to provide DP guarantee. Also, it employs homomorphic encryption such that the teacher models can perform the computation on an encrypted query, which prevents the teachers from viewing the raw input of the querying party. One can use the teachers to label (with DP) a public dataset or their own private dataset (that is large enough) to train a student model for their own usage. However, having access to a public dataset that has negligible privacy concerns is not always a realistic assumption in the healthcare setting. In addition, in order for PATE or CaPC to achieve good performance while maintaining a reasonable privacy guarantee, it usually requires a large number of participants, which may be hard to achieve for the healthcare settings. Hence, for our approach, we decide to follow the general category of merging model updates, instead of using DP labels from the ensemble of local models.

By thoroughly examining existing collaborative learning frameworks, we have identified that the threat model is not clearly defined for the needs of hospitals. As a result, current frameworks lack some crucial components needed to enable multiple hospitals to collaborate and conduct large-scale ML training in a secure and private manner. In our research, we aim to address this gap by clearly defining the threat model for collaborations among hospitals and selecting appropriate tools to ensure the privacy of patients' data. Furthermore, we will propose a new collaborative ML framework that will provide a secure and private solution for healthcare organizations to collaborate and train ML models on their distributed datasets without compromising patient privacy.

\section*{Membership Inference Attack}
\label{supp_sec:mia_background}
Membership inference attack (MIA)~\cite{shokri2017membership,carlini2022membership} is a commonly used attack to evaluate how much information could leak from a machine learning model. The goal of the attack is to infer whether a data point is included in the training dataset of an ML model or not. It can be used as an attack or a way to empirically evaluate the privacy bound of a model. There are many proposed implementations of MIA. Most of them analyse the model’s outputs (e.g., loss values or the model’s confidence score): intuitively, if a data point exists in the training dataset, the model tends to have a lower loss value and higher confidence score. An adversary then queries the target model about the data points, gets the loss value or confidence score, and classifies if the point is a member or a non-member. In this paper, we evaluate the robustness of the models using Likelihood Ratio Attack (LiRA)~\cite{carlini2022membership}, one of the strong MIAs.

In our experiments, we assume the strongest adversary who has access to the model architecture, true data distribution, and all the hyperparameters used to train the model, which means the entire training process. This assumption is reasonable since our goal is to evaluate/audit the amount of information leaked about the training data points, rather than performing an actual attack. We use the Online version of LiRA introduced in the original paper. For the target models, we randomly partition the entire dataset into two disjoint subsets, and one of them is used to train the models. This training subset forms the members of the target models, and the other subset forms the non-members. 

In order to audit how vulnerable the membership information is for the target models, the adversary uses the same model architecture and training process as the target models to train several shadow models. Each of the shadow models is trained with half of the dataset as its members and the other half forms its non-members. Therefore, for each data point in the dataset, it appears as a member for half of the shadow models and as a non-member for the other half. The adversary queries all the shadow models with each of the data points to obtain the confidence scores based on the predictions of the shadow models. These scores are then used to fit a distribution for such model architecture and training process about how confident the model would be when the data point is a member and vice versa. Next, we query the target models with the data points for their confidence scores and compare the scores with the distribution estimated from the shadow models’ performance to predict if the data point is a member of the training data for the target model. By evaluating this prediction against the ground truth values of membership information for the target model, we plot the True Positive Rate (TPR) vs. False Positive Rate (FPR) of the adversary’s prediction at various thresholds, also known as the ROC curve.

\section*{Experimental Setup}
\label{supp_sec:experimental_setup}
In order to train an ML model to be differentially private, the state-of-art algorithm is DP-SGD~\cite{Abadi2016dpsgd}, which is also used by PriMIA and our \ours framework. In this work, we use Opacus~\cite{opacus} version 0.15.0 for all our experiments to evaluate our framework and PriMIA. 
For GEMINI~\cite{Verma2017gemini,Verma2021_gemini}, we use $C=1.0$ for training with the MLP model and $C=0.5$ the Logistic regression. We use a privacy budget of $\epsilon=2.0$ for both PriMIA and our framework.
For Pancreas dataset, we use $C=0.5$ and $\sigma=1.0$ for both MLP and SVC model architectures. The privacy budget of $\epsilon=5.6$ is used when training using our framework and PriMIA. 
For X-ray, we use $C=0.5$ and $\sigma=1.0$ for both settings pre-trained~\cite{pan2010transfer_learning} with MIMIC-CXR~\cite{Johnson2019mimic,Johnson2019mimic-physionet,Goldberger2000mimic-physionet} and ImageNet~\cite{deng2009imagenet}. In terms of the privacy budget, we use $\epsilon=0.62$ and $\epsilon=0.65$ for MIMIC-CXR and ImageNet pre-training cases respectively. For all the experiments, we use $\delta=\min\{10^{-5}, 1.1 \times \text{size of dataset}\}$, where for PriMIA, the dataset refers to the individual private training dataset since PriMIA protects local DP; for our \ours framework, the size of the dataset refers to the summation of all private training datasets since we try to achieve distributed DP. The order ($\alpha$) of RDP~\cite{Mironov2017RDP} is the optimum value calculated by Opacus).

\subsection*{GEMINI study}
For MLP, we use $\eta=0.01$ for all non-DP training and $\eta=0.1$ for all DP training. The models are training for $100$ epochs or until the privacy budgets are used up. 
For Logistic Regression, learning rate $\eta=0.15$ for all experiments. The models are trained for $30$ epochs or until the privacy budgets are used up. To prevent overfitting, we apply standard $l_2$ regularization (weight decay of $0.0002$). 
For the MIA experiments, we use the same setup as previous experiments except that we set $\epsilon$ to be $9.0$, allowing the model to train for longer and converge better on the smaller training dataset.  We use only the original data point (no additional augmentations) to calculate the LiRA scores. 

\subsection*{Pancreas Study}
To prevent overfitting, we apply standard $l_2$ regularization (weight decay of 0.0002). For MLP, we use learning rate $\eta=0.03$ for non-DP training (FL and training with individual private datasets). We use $\eta=0.1$ when training with our framework and PriMIA. For SVC, we use $\eta=0.1$ for all experiments. We trained the models for 50 epochs or until when the privacy budget $\epsilon=5.6$ is used up. 
For the MIA experiments, we use the same setup as previously discussed except that we use $\epsilon=9.0$. Using a larger privacy budget would allow the model to converge. We use only the original data point (no additional augmentations) to calculate the LiRA scores. 

\subsection*{Chest Radiology}
For reference, the pre-trained model’s performance on the 5-fold test datasets are summarised in Supplementary Table~\ref{tab:pretrained_utility} with mean AUROC and the standard deviation.

For all the experiments, we use a weight decay of $1.0\times10^{-5}$. For all non-DP training (FL and training with individual private datasets), the learning rate $\eta=0.001$. For PriMIA and our \ours framework, we use $\eta=0.01$. For consistency, we freeze the batch normalization layer of the model for all our experiments. For the experiments with model pre-trained on MIMIC-CXR, we use a privacy budget of $\epsilon=0.62$. For FL, PriMIA, and our \ours framework, we train the models for $3$ epochs or until the target privacy budget is reached. For the experiments with models pre-trained on ImageNet (provided by TorchVision), since the model weights provided are pre-trained on coloured images with 3 input channels but the chest radiology images have only 1 channel (grey-scale image), we take the summation of the weights of the 3 channels for the input convolutional layer. We use a privacy budget of $0.65$. For FL, PriMIA, and \ours framework, we train the models for $7$ epochs or until the target privacy budget is reached. In both pre-training cases for the models trained with only one of the private datasets, we train the models for 10 epochs.
For LiRA experiments, since LiRA doesn’t support multilabel classification models, we slightly modified the training setting: we only filter for images with frontal views (``AP" and ``PA") but we don’t filter for the most common three pathologies; also, we use DenseNet121 with only one output neuron which only predicts if the input has ``No Finding" or not). The class distributions used for LiRA experiments are shown in Supplementary Table~\ref{tab:xray_lira_dataset_stats}. In this setting, we use $\eta=0.005$ and $\eta=0.01$ for non-DP and DP training respectively.

We train the target models and shadow models using the half of each dataset as the training dataset. We train 64 shadow models. We also train 5 target models both with DP and without DP to get the confidence interval. We use 4 augmentations of each data point to compute the score for LiRA to determine whether the point is a member of not. All models are trained for 60 epochs or until the privacy budget of $\epsilon=1.0$. 

\section*{Experimental Results}
\label{supp_sec:additional_exp_results}
\subsection*{\ours predicts mortality of patients admitted to hospitals using EHR}
In this study, we investigate the effectiveness of \ours for the mortality prediction task using electronic health records (EHRs) collected from the GEMINI initiative~\cite{Verma2017gemini,Verma2021_gemini}. We use two types of models, multilayer perceptron (MLP) models and linear models to run regression on the dataset. The results for the MLP regressor are summarised in Supplementary Table~\ref{tab:results_gemini_mlp} and main Figure 2c. The results of using the linear model (i.e., logistic regression analysis) for this task are summarised in Supplementary Figure~\ref{fig:results_gemini_logReg} and Supplementary Table~\ref{tab:results_gemini_logReg}. For both model architectures, FL and \ours lead to improved model performance compared to models trained with only one private dataset. Moreover, our framework allows us to train models with a privacy guarantee while having little utility trade-off compared to the models trained with FL. 

We also note that linear models are more robust to privacy attacks than MLP models when both of them are trained without privacy guarantees, as shown in Supplementary Figure~\ref{fig:mia_gemini_logReg}. The AUROC scores for the membership inference attack on the two target linear models (trained with FL and \ours framework) are similar regardless of whether they are trained with or without privacy guarantee. 
In addition, as we compare this with the main Figure 5a, we observe that non-DP linear models are much more robust than the non-DP MLP models. This observation suggests that using a simpler model architecture (i.e., the one with fewer parameters) may reduce the risk of privacy leakage when different model architectures have similar performance. Finally, our models trained with \ours framework outperform the models trained with PriMIA when using the same privacy budget.

\subsection*{\ours classifies cell types in single-cell human pancreas studies}
In this study, we explore the effectiveness of \ours on cell classification tasks using single-cell human pancreas datasets. We employ two types of classifiers, namely MLP and Support Vector Classifier (SVC), for this analysis. The summary of results for the MLP classifier is presented in Supplementary Table~\ref{tab:results_pancreas_mlp} and main Figure 3c. Additionally, we present the results of using SVC for the same task in Supplementary Figure~\ref{fig:results_pancreas_svc} and Supplementary Table~\ref{tab:results_pancreas_svc}. 
Our findings indicate that both FL and \ours lead to improved utility compared to models trained with only one private dataset available at each silo. However, our framework allows us to train models with a privacy guarantee while maintaining high utility with little utility trade-off compared to the models trained with FL. Finally, our models trained with \ours framework outperform the models trained with PriMIA when using the same privacy budget. 

In addition, we conduct membership inference attack on SVC target models for this task. The results are summarised in Supplementary Figure~\ref{fig:mia_pancreas_svc}. For this task, the trend is similar to that of MLP model architecture: the models trained with FL without DP guarantee is much more vulnerable to privacy attacks than the models trained with \ours. This phenomenon is different from the scenario where a linear model is used to run a logistic regression analysis on the GEMINI dataset. In that case, linear models trained with and without DP guarantee are almost equally robust under LiRA. This may be because the SVC has the capacity to learn the dataset well enough like the MLP model.

\subsection*{\ours identifies pathologies from human chest radiology images}
In this study, we evaluate the effectiveness of \ours for the pathology identification task using human chest radiology images. We employ the transfer learning technique and model states pre-trained on MIMIC-CXR and ImageNet. The summary of the results for the scenario where the initial model is pre-trained on MIMIC-CXR is provided in main Figure~4c and Supplementary Table~\ref{tab:results_xray_mimic}. 
We run additional experiments on the setting where the initial state of the model is pre-trained on ImageNet. The results are summarized in Supplementary Figure~\ref{fig:results_xray_imgnet} and Supplementary Table~\ref{tab:results_xray_imgnet}.

\section*{Data Collection}
\label{supp_sec:data_collection}
\subsection*{GEMINI}
\label{supp_ssec: data_collection_gemini}
We gather the health records of all patients admitted from April 1, 2015, to January 23, 2020 from the GEMINI dataset~\cite{Verma2017gemini,Verma2021_gemini}. The cohort includes the patients admitted to any medical or ICU service (except cardiology). This dataset was created to predict hospital mortality in non-cardiac patients, and thus the {the following patients were excluded}: patients admitted to cardiology or coronary ICU, patients with {Main Emergency Department diagnosis} being a cardiac condition (CAD, CHF, arrhythmia), patients with troponin elevated before admission and up to the first 24 hours after admission, patients who are discharged (or dead) before 24 hours after admission. We only include the most recent visit of each patient and we keep track of the total number of past visits for each patient as a feature. Each patient is identified based on health insurance number in the database. Hence we filter the patients that do not have a recorded health insurance number. 
Some of the admission information is included as input features in our analysis, for example, the age of the patients at the admission date, the date of admission (day of the week and month), readmission category, and whether the patient is transferred from acute care. We also include the mode of admission and triage level. 

We consider the following imaging tests: X-ray, CT, MRI, and US. For each of the imaging tests, we consider the following body parts head, neck, chest, breast, abdomen, pelvis, limb, shoulder, and the whole body. In addition, we include echocardiography. 

For blood transfusion, we consider the following four products: red blood cells, plasma, platelet, and albumin. 

Note that it's important for participating hospitals to standardize the name of the features, the categories of the features if categorical, and the units of the features if numerical. For the ML approaches we are using in our analysis, this is an important step. If the categories or the feature names are not standardized, we may end up with too many one-hot encoded features. In addition, if the features are not recorded in the same unit, these two things would prevent the models from learning and making meaningful predictions. 

Therefore, the participating hospitals should communicate about all of this information (e.g., admission categories, available tests) and standardize the names, categories, and units. {This standardization is performed after data are collected from hospitals in GEMINI.} This will not result in any significant privacy leakage comparing the sharing of model weights/gradient updates during the training. This does not violate the purpose of evaluating the performance of the framework on cross-silo datasets.

In summary, all the categorical features and numerical features are summarised in Supplementary Table~\ref{tab:cat_features_gemini} and Supplementary Table~\ref{tab:num_features_gemini} respectively. Note that different laboratories may use different assays, thus the laboratory test results (used as numerical features) may not be fully standardized in this study.

\subsection*{Pancreas}
For the single-cell human pancreas datasets, we use the same processed version used by~\cite{Wang2022ocat}, which can be downloaded from \url{https://data.wanglab.ml/OCAT/Pancreas.zip}.

\subsection*{Chest radiology}
The X-ray datasets are collected from NIH~\cite{Wang2017nih-chestx-ray8}, PadChest (PC)~\cite{bustos_padchest_2020}, CheXpert (CheX)~\cite{irvin2019chexpert}. For pretraining, we also use MIIMIC-CXR~\cite{Johnson2019mimic, Johnson2019mimic-physionet, Goldberger2000mimic-physionet}. 
NIH contains 112,120 images collected from 30,805 patients. The data can be downloaded from \url{https://www.nih.gov/news-events/news-releases/nih-clinical-center-provides-one-largest-publicly-available-chest-x-ray-datasets-scientific-community}. We use a downsized version pre-processed by TorchXrayVision~\cite{cohen2020torchxrayvision1,cohen2022torchxrayvision2} which can be downloaded from \url{https://academictorrents.com/details/e615d3aebce373f1dc8bd9d11064da55bdadede0}. 
PC contains 160,000 images from 67,000 patients. The data can be downloaded from \url{https://bimcv.cipf.es/bimcv-projects/padchest/}. We use a downsized version pre-processed by TorchXrayVision, which can be downloaded from \url{https://academictorrents.com/details/96ebb4f92b85929eadfb16761f310a6d04105797}. 
CheX contains 224,316 chest radiographs of 65,240 patients. The data can be downloaded from \url{https://stanfordmlgroup.github.io/competitions/chexpert/}. 
MIMIC-CXR contains 377,110 images of 65,379 patients. The data can be downloaded from \url{https://physionet.org/content/mimic-cxr-jpg/2.0.0/}. 
For all four datasets, we only include the images with view position = [``AP", ``PA"] (frontal view) for our experiments. The output labels for analysis are the 3 most common pathologies that appeared in the four for-mentioned datasets (``Atelectasis'', ``Effusion'', and ``Cardiomegaly''). Data points with ``No Finding" are also included in the analysis as the control.  
After filtering, the number of cases for each dataset is summarised in Supplementary Table~\ref{tab:xray_dataset_stats} and we use the filtered datasets to conduct the case study (as well as pre-train the model on MIMIC-CXR).

\section*{Communication Cost of the Protocol}
\label{supp_sec:sec_agg_and_comm_cost}
Secure Aggregation (SecAgg)~\cite{Bonawitz2017SecAgg, Bell2020SecAggPlus} is a class of Secure Multiparty Computation where a group of participants can securely compute the summation over their private values without knowing the values of others. We leverage this tool in our protocol such that when the leader performs the aggregation, it will not see other participants’ model updates but an aggregate sum. This sum has been clipped and noised, hence the intermediate model states are all differentially private (more about DP in the Online Method Section). 

In Supplementary Figure~\ref{fig:sec_agg_asympt}, we use SecAgg introduced by ~\cite{Bonawitz2017SecAgg} to empirically test the wall clock time needed for different numbers of clients and different model sizes. We record the time from starting the server until the server finishes calculating the weights.

To analyse the additional cost associated with SecAgg when aggregating the model updates in our previous case studies, we conduct a comparison of the communication (in megabytes) for the participants and the leader in scenarios with and without SecAgg per communication round. We use the same model architectures as previous experiments. 
The experimental results are summarised in Supplementary Table~\ref{tab:communication_time}. The computation overheads for the participants and the aggregator are minor (less than one second) for all the case studies evaluated in this work, hence we do not report them separately.

Note that the state of the art secure aggregation can reach (poly)logarithmic complexity~\cite{Bell2020SecAggPlus}. Because there is no source code provided for this technique, we did not evaluate it here.

\clearpage
\section*{Supplementary figures}
\begin{figure}[h]
\centering
\subfloat[Time vs. Num Clients]{\includegraphics[width=0.5\linewidth]{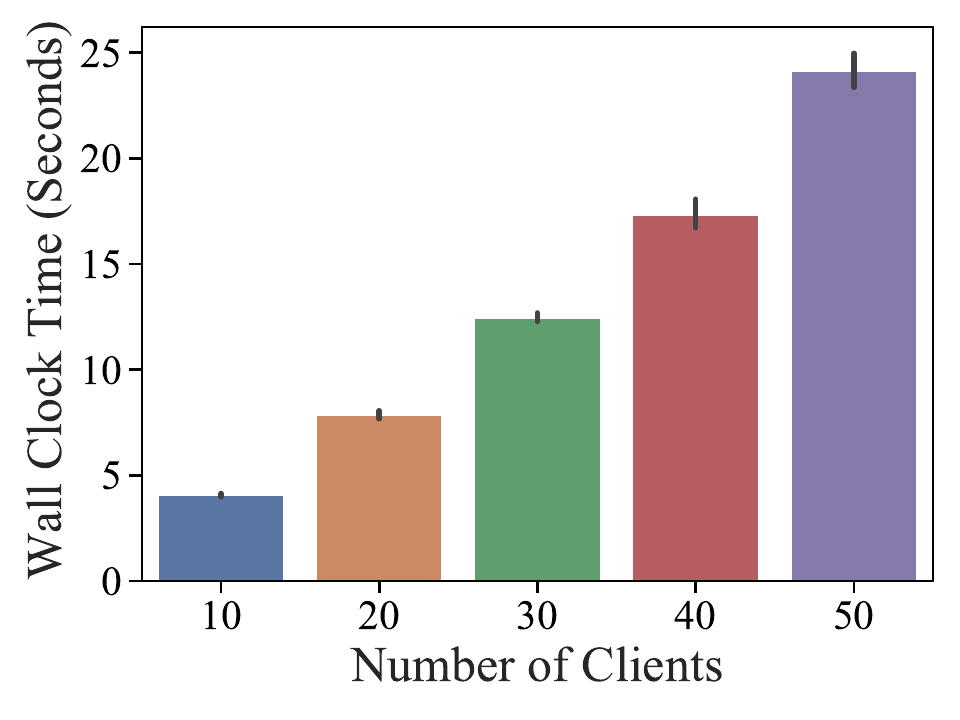}}
\subfloat[Time vs. Input Dim]{\includegraphics[width=0.5\linewidth]{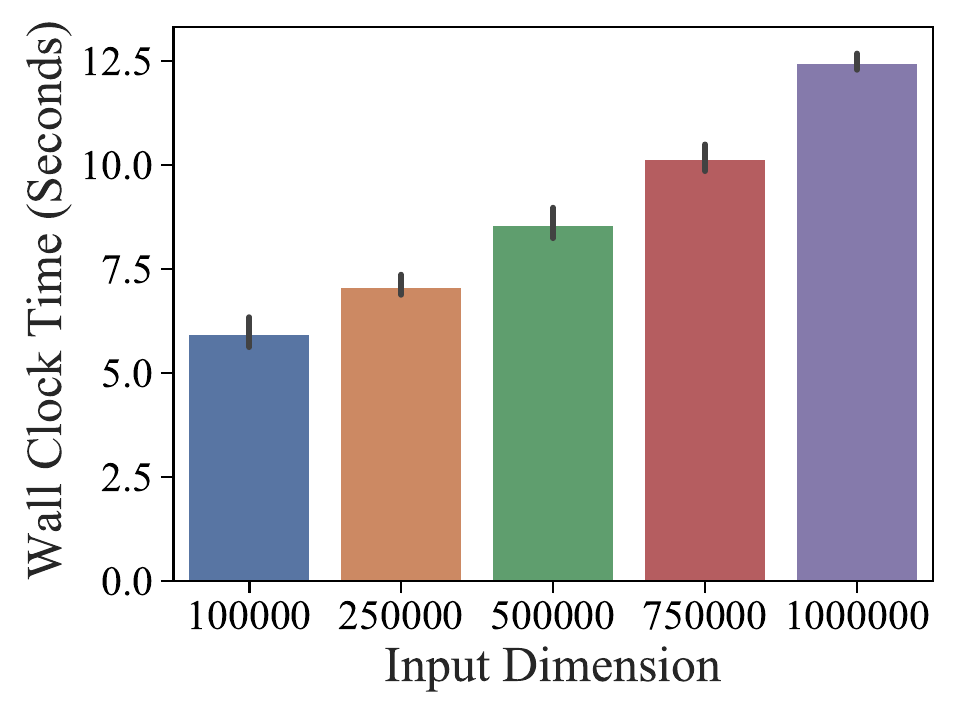}}
\caption{ \textbf{Evaluation of total wall-clock time for SecAgg per communication round.} The clients and server are simulated on the same machine with 32 cores and 64G memory. The experiments are repeated 5 times and we report the $95\%$ confidence interval. (\textbf{a}), input dimension is fixed to be 1,000,000 and the number of participants/clients varied. (\textbf{b}), number of clients is fixed to be 30 and the input dimension is varied. }
\label{fig:sec_agg_asympt}
\end{figure}

\begin{figure}[h]
\centering\hspace{-5mm}
\includegraphics[width=\linewidth]{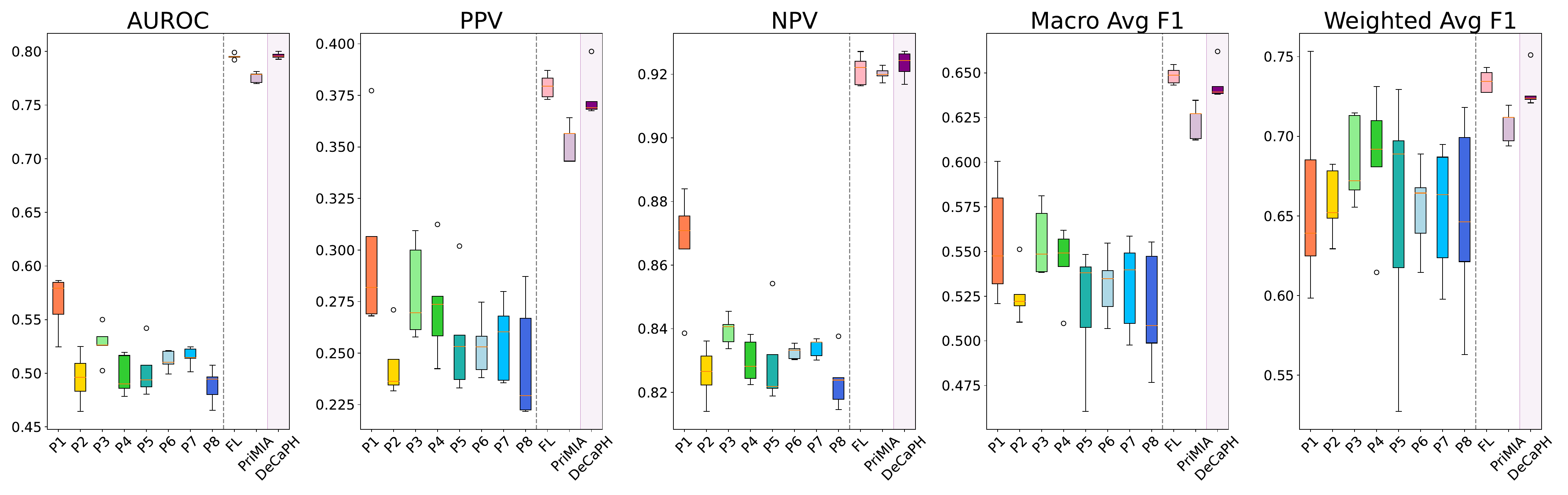}
\caption{ \textbf{Comparison of models trained by different collaborative training frameworks to predict patient mortality.} 
{The figures show the first quartile, median, and third quartile, as well as the outliers ($1.5 \times$ interquartile range below or above the lower and upper quartile.) We perform a Wilcoxon signed-rank test (one-
tail) with continuity correction using exact method to compare the performance of models trained with \ours to those
trained with PriMIA for each of the evaluation metrics. The alternative hypothesis is that models trained with \ours have higher scores. The p-values are $< 0.05$ for all metrics except for the NPV metric. }
}
\label{fig:results_gemini_logReg}
\end{figure}

\begin{figure}[h]
\centering\hspace{-5mm}
\includegraphics[width=\linewidth]{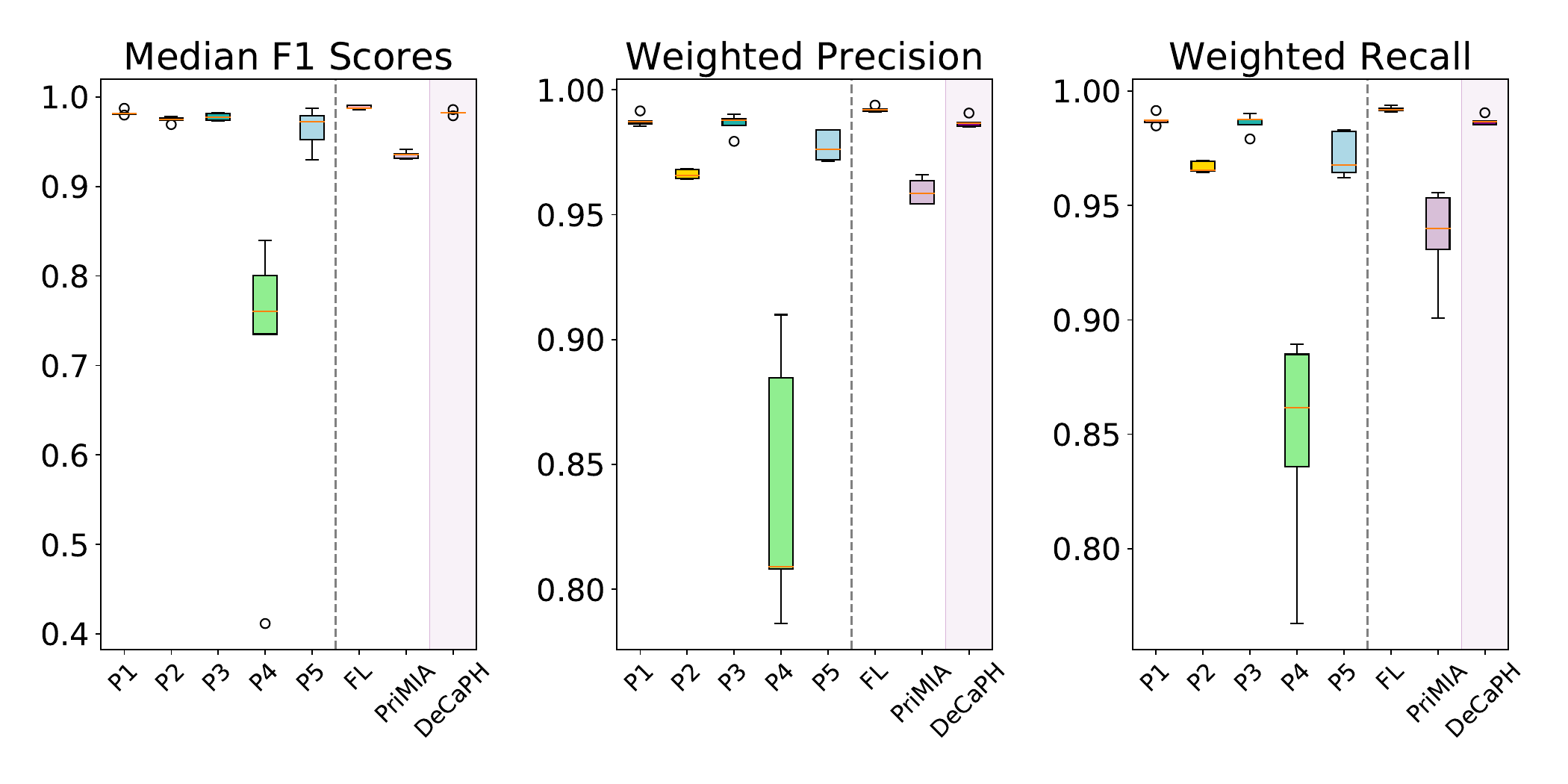}
\caption{ \textbf{Comparison of models trained by different collaborative training frameworks to classify cell types. }{The figures show the first quartile, median, and third quartile, as well as the outliers ($1.5 \times$ interquartile range below or above the lower and upper quartile.) We perform a Wilcoxon signed-rank test (one-tail) with continuity correction using exact method on performance
of models trained with \ours and PriMIA for each of the evaluation metrics. The alternative hypothesis is that models
trained with \ours have higher scores for that metric. The p-values are $<0.05$ for all metrics.} }
\label{fig:results_pancreas_svc}
\end{figure}

\begin{figure}[h]
\centering\hspace{-5mm}
\includegraphics[width=\linewidth]{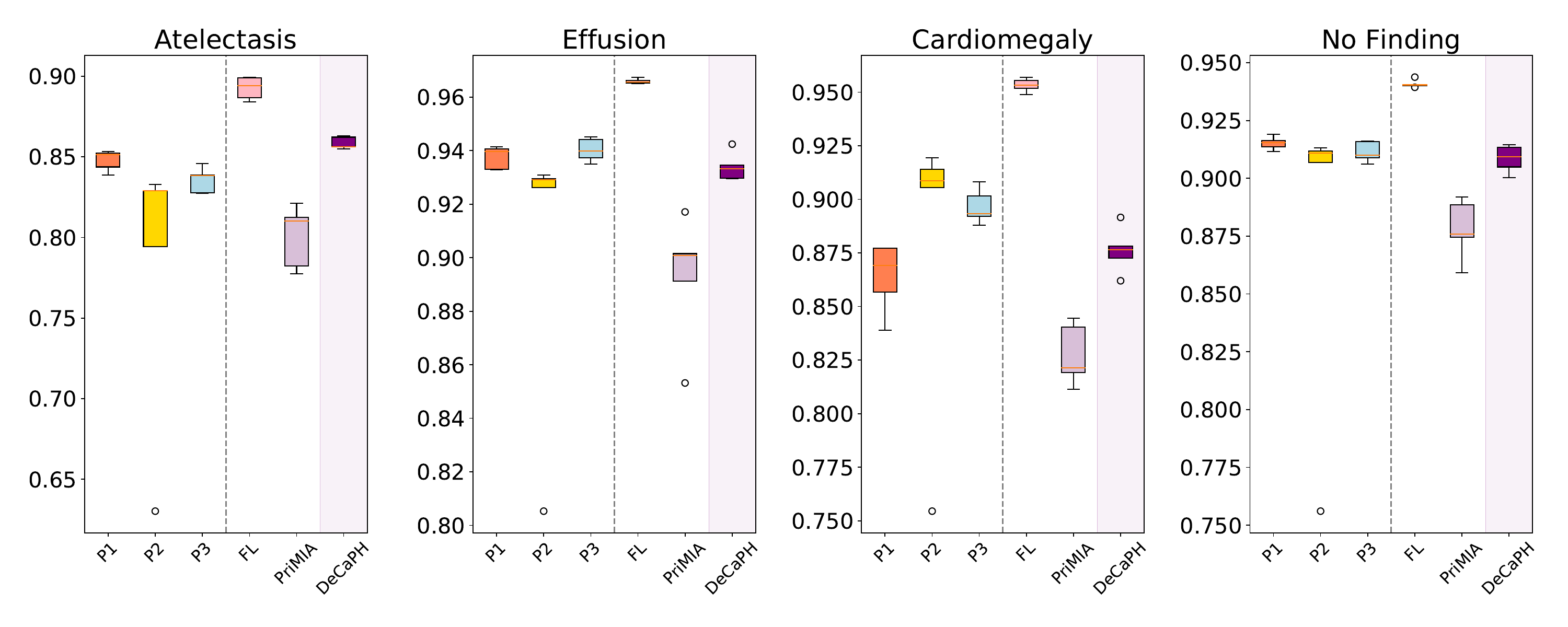}
\caption{ \textbf{Comparison of models trained by different collaborative training frameworks to identify pathologies.} {The figures show the first quartile, median, and third quartile, as well as the outliers ($1.5 \times$ interquartile range below or above the lower and upper quartile.) We perform a Wilcoxon signed-rank test (one-tail) with continuity correction using exact method on performance of models trained with \ours and PriMIA for each of the pathologies and ``No Finding". The alternative hypothesis is that models trained with \ours have higher AUROC scores. The p-values are $< 0.05$ for all three pathologies and ``No Finding". }}
\label{fig:results_xray_imgnet}
\end{figure}

\begin{figure}[h]
\centering\hspace{-5mm}
\includegraphics[width=0.5\linewidth]{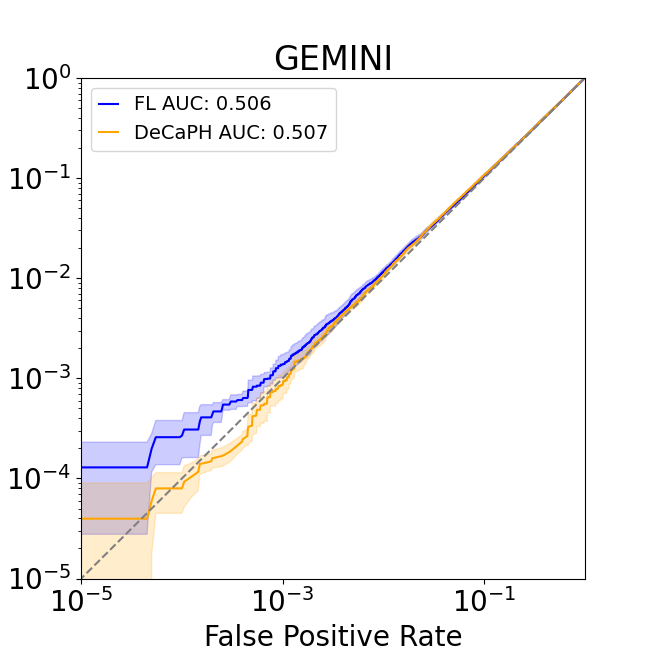}
\caption{\textbf{Membership Inference Attack on linear models trained with DP vs. without DP for patient mortality prediction task using GEMINI EHR datasets.} The target models are trained for five times to plot the $95\%$ confidence interval. AUROC for FL is  $0.506 \pm 0.001$ and that for \ours (Ours) is $0.507 \pm 0.003${; mean $\pm$ SD.} }
\label{fig:mia_gemini_logReg}
\end{figure}

\begin{figure}[h]
\centering\hspace{-5mm}
\includegraphics[width=0.5\linewidth]{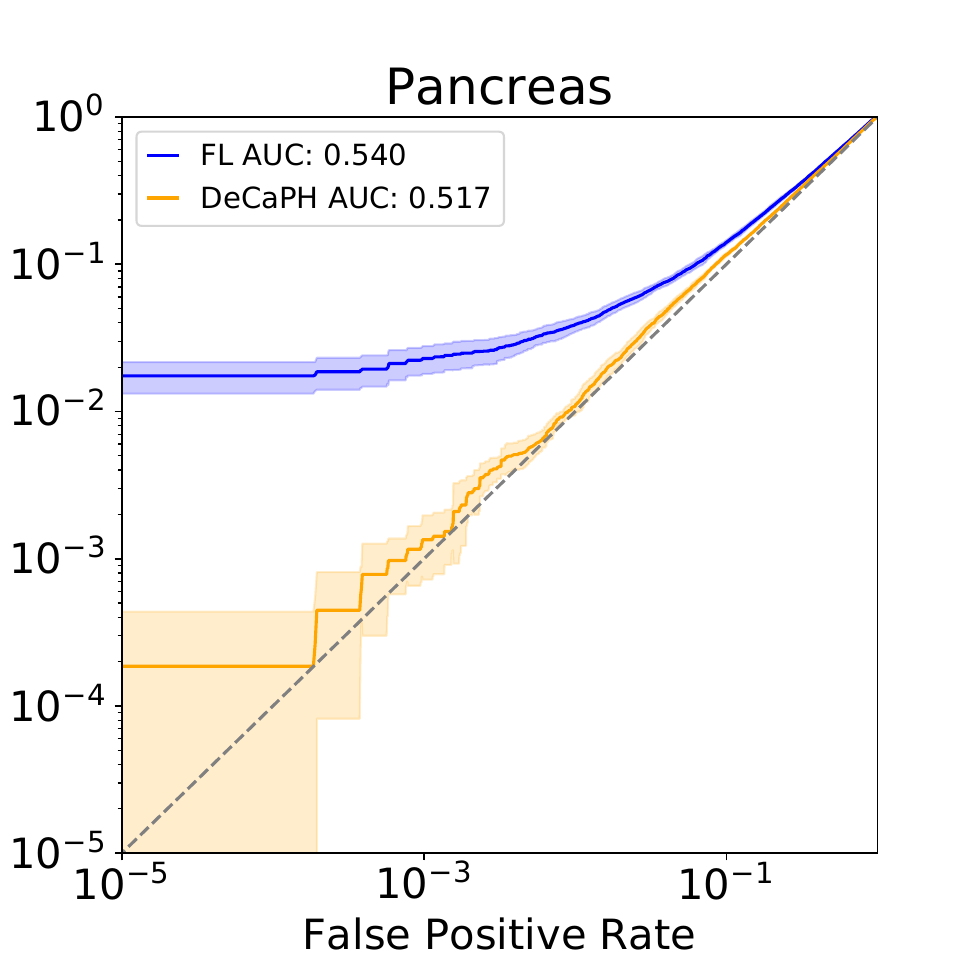}
\caption{\textbf{Membership Inference Attack on SVC models trained with DP vs. without DP for cell classification task using single-cell human pancreas dataset.} The target models are trained for five times to plot the $95\%$ confidence interval. AUROC for FL is  $0.540 \pm 0.003$ and that for \ours (Ours) is $0.517 \pm 0.004${; mean $\pm$ SD.} }
\label{fig:mia_pancreas_svc}
\end{figure}

\clearpage
\section*{Supplementary tables}
\begin{table}[h]
\centering
\normalsize
\begin{tabular}{|l|l|l|l|l|l|}
\hline \multirow{3}{*}{$\begin{array}{l}\text { Task (\# } \\ \text { participants) }\end{array}$} & \multirow{3}{*}{$\begin{array}{l}\text {\# param } \\\text { (archt.) }\end{array}$} & \multicolumn{4}{c|}{ Communication Cost (MB) } \\
\cline { 3 - 6 } & & \multicolumn{2}{l}{With SecAgg} & \multicolumn{2}{|l|}{ Without SecAgg } \\
\cline { 3 - 6 } & & Per Participant & Aggregator & Per Participant & Aggregator \\
\hline GEMINI (8) & $\begin{array}{l}166,771 \\
\text { (MLP) }\end{array}$ & 3257.316 & 26058.531 & 1302.898 & 10423.188 \\
\cline { 2 - 6 } & $\begin{array}{l}437 \\
\text { (Linear) }\end{array}$ & 8.605 & 68.844 & 3.414 & 27.312 \\
\hline Pancreas (5) & $\begin{array}{l}15,659,504 \\
\text { (MLP) }\end{array}$ & 305849.734 & 1529248.672 & 122339.875 & 611699.375 \\
\cline { 2 - 6 } & $\begin{array}{l}62,236 \\
\text { (Linear) }\end{array}$ & 1215.594 & 6077.969 & 486.219 & 2431.094 \\
\hline X-ray (3) & $\begin{array}{l}7,035,453 \\
\text { (DenseNet) }\end{array}$ & 137411.223 & 412233.668 & 54964.477 & 164893.430 \\
\hline
\end{tabular}
\caption{ \textbf{Average communication cost with and without SecAgg per communication round}. }
\label{tab:communication_time}
\end{table}

\begin{table}[h]
    \centering
    \begin{tabularx}{\textwidth}{|X|l|l|l|l|}
\hline  \textbf{Pre-trained Dataset} & \textbf{Atelectasis} & \textbf{Effusion} & \textbf{Cardiomegaly} & \textbf{No Finding} \\
\hline \textbf{MIMIC-CXR} & $0.8475 \pm 0.0031$ & $0.9371 \pm 0.0005$ & $0.9013 \pm 0.0030$ & $0.9071 \pm 0.0010$ \\
\hline \textbf{ImageNet} & $0.5449 \pm 0.1409$ & $0.5106 \pm 0.0748$ & $0.4501 \pm 0.1075$ & $0.5567 \pm 0.1711$ \\
\hline
\end{tabularx}
    \caption{\textbf{The utilities of the pre-trained models}{; mean $\pm$ SD.}}
    \label{tab:pretrained_utility}
\end{table}

\begin{table}[h]
    \centering
    \begin{tabular}{|l|l|l|l|}
\hline & \textbf{Labels} & \textbf{No Finding} & \textbf{Total Number} \\
\hline \multirow{2}{*}{ \textbf{NIH} } & 1 & 60361 & 112120 \\
\cline { 2 - 3 } & 0 & 51759 & \\
\hline \multirow{2}{*}{ \textbf{PC} } & 1 & 48963 & 94825 \\
\cline { 2 - 3 } & 0 & 45862 & \\
\hline \multirow{2}{*}{ \textbf{CheX} } & 1 & 16970 & 191010 \\
\cline { 2 - 3 } & 0 & 174040 & \\
\hline
\end{tabular}
    \caption{\textbf{Number of datapoints of the Chest Radiology datasets used for LiRA experiments}}
    \label{tab:xray_lira_dataset_stats}
\end{table}

\begin{table}[h]
    \centering
    \begin{tabular}{|l|c|c|c|c|c|}
\hline  & \textbf{AUROC} & \textbf{PPV} & \textbf{NPV} & \textbf{Macro Avg F1} & \textbf{Weighted Avg F1} \\
\hline \textbf{Hospital 1 (P1)} & $0.540 \pm 0.021$ & $0.343 \pm 0.030$ & $0.846 \pm 0.009$ & $0.593 \pm 0.008$ & $0.732 \pm 0.017$ \\
\hline \textbf{Hospital 2 (P2)} & $0.487 \pm 0.018$ & $0.305 \pm 0.028$ & $0.824 \pm 0.005$ & $0.559 \pm 0.013$ & $0.727 \pm 0.011$ \\
\hline \textbf{Hospital 3 (P3)} & $0.516 \pm 0.060$ & $0.242 \pm 0.045$ & $0.815 \pm 0.054$ & $0.491 \pm 0.084$ & $0.601 \pm 0.119$ \\
\hline \textbf{Hospital 4 (P4)} & $0.515 \pm 0.055$ & $0.308 \pm 0.049$ & $0.837 \pm 0.020$ & $0.554 \pm 0.017$ & $0.702 \pm 0.044$ \\
\hline \textbf{Hospital 5 (P5)} & $0.478 \pm 0.048$ & $0.222 \pm 0.059$ & $0.806 \pm 0.031$ & $0.471 \pm 0.045$ & $0.606 \pm 0.077$ \\
\hline \textbf{Hospital 6 (P6) }& $0.520 \pm 0.064$ & $0.233 \pm 0.037$ & $0.823 \pm 0.044$ & $0.487 \pm 0.062$ & $0.597 \pm 0.093$ \\
\hline \textbf{Hospital 7 (P7)} & $0.502 \pm 0.054$ & $0.239 \pm 0.043$ & $0.816 \pm 0.035$ & $0.501 \pm 0.062$ & $0.626 \pm 0.087$ \\
\hline \textbf{Hospital 8 (P8)} & $0.547 \pm 0.042$ & $0.239 \pm 0.008$ & $0.861 \pm 0.035$ & $0.467 \pm 0.041$ & $0.549 \pm 0.095$ \\
\hline \textbf{FL} & $0.822 \pm 0.002$ & $0.402 \pm 0.017$ & $0.933 \pm 0.007$ & $0.669 \pm 0.014$ & $0.751 \pm 0.017$ \\
\hline \textbf{PriMIA} & $0.776 \pm 0.003$ & $0.352 \pm 0.005$ & $0.920 \pm 0.004$ & $0.622 \pm 0.005$ & $0.706 \pm 0.007$ \\
\hline \textbf{\ours (Ours)} & $0.817 \pm 0.003$ & $0.416 \pm 0.013$ & $0.926 \pm 0.007$ & $0.678 \pm 0.009$ & $0.763 \pm 0.012$ \\
\hline
\end{tabular}
    \caption{\textbf{The utilities of models with MLP architecture for patient mortality prediction.} The hospital names are anonymized; mean $\pm$ SD.}
    \label{tab:results_gemini_mlp}
\end{table}

\begin{table}[h]
    \centering
        \begin{tabular}{|l|c|c|c|c|c|}
\hline  & \textbf{AUROC} & \textbf{PPV} & \textbf{NPV} & \textbf{Macro Avg F1} & \textbf{Weighted Avg F1} \\
\hline \textbf{Hospital 1 (P1)} & $0.566 \pm 0.024$ & $0.301 \pm 0.041$ & $0.867 \pm 0.015$ & $0.556 \pm 0.030$ & $0.660 \pm 0.054$ \\
\hline \textbf{Hospital 2 (P2)} & $0.496 \pm 0.021$ & $0.244 \pm 0.014$ & $0.826 \pm 0.008$ & $0.526 \pm 0.014$ & $0.658 \pm 0.020$ \\
\hline \textbf{Hospital 3 (P3)} & $0.528 \pm 0.015$ & $0.280 \pm 0.021$ & $0.839 \pm 0.004$ & $0.556 \pm 0.018$ & $0.684 \pm 0.025$ \\
\hline \textbf{Hospital 4 (P4)} & $0.498 \pm 0.017$ & $0.273 \pm 0.023$ & $0.830 \pm 0.006$ & $0.544 \pm 0.018$ & $0.686 \pm 0.039$ \\
\hline \textbf{Hospital 5 (P5)} & $0.502 \pm 0.022$ & $0.257 \pm 0.025$ & $0.830 \pm 0.013$ & $0.519 \pm 0.032$ & $0.652 \pm 0.072$ \\
\hline \textbf{Hospital 6 (P6)} & $0.512 \pm 0.008$ & $0.253 \pm 0.013$ & $0.833 \pm 0.002$ & $0.531 \pm 0.017$ & $0.655 \pm 0.026$ \\
\hline \textbf{Hospital 7 (P7)} & $0.515 \pm 0.008$ & $0.256 \pm 0.017$ & $0.834 \pm 0.003$ & $0.531 \pm 0.023$ & $0.653 \pm 0.037$ \\
\hline \textbf{Hospital 8 (P8)} & $0.489 \pm 0.015$ & $0.245 \pm 0.027$ & $0.824 \pm 0.008$ & $0.517 \pm 0.030$ & $0.650 \pm 0.056$ \\
\hline \textbf{FL} & $0.795 \pm 0.002$ & $0.379 \pm 0.005$ & $0.921 \pm 0.004$ & $0.648 \pm 0.004$ & $0.735 \pm 0.006$ \\
\hline \textbf{PriMIA} & $0.776 \pm 0.004$ & $0.353 \pm 0.008$ & $0.920 \pm 0.002$ & $0.623 \pm 0.009$ & $0.707 \pm 0.010$ \\
\hline \textbf{\ours (Ours)} & $0.796 \pm 0.002$ & $0.375 \pm 0.011$ & $0.923 \pm 0.004$ & $0.644 \pm 0.009$ & $0.729 \pm 0.011$ \\
\hline
\end{tabular}
    \caption{\textbf{The utilities of Logistic Regression models for patient mortality prediction.} The hospital names are anonymized{; mean $\pm$ SD.}}
    \label{tab:results_gemini_logReg}
\end{table}

\begin{table}[h]
    \centering
    \begin{tabular}{|l|c|c|c|}
\hline & \textbf{Median F1 Scores} & \textbf{Weighted Precision} & \textbf{Weighted Recall} \\
\hline \textbf{Baron (P1)} & $0.983 \pm 0.002$ & $0.987 \pm 0.002$ & $0.986 \pm 0.002$ \\
\hline \textbf{Muraro (P2)} & $0.986 \pm 0.004$ & $0.985 \pm 0.003$ & $0.985 \pm 0.003$ \\
\hline \textbf{Seg (P3)} & $0.985 \pm 0.002$ & $0.991 \pm 0.001$ & $0.991 \pm 0.001$ \\
\hline \textbf{Wang (P4)} & $0.534 \pm 0.106$ & $0.784 \pm 0.046$ & $0.822 \pm 0.026$ \\
\hline \textbf{Xin (P5)} & $0.972 \pm 0.004$ & $0.985 \pm 0.002$ & $0.984 \pm 0.003$ \\
\hline \textbf{FL} & $0.989 \pm 0.003$ & $0.993 \pm 0.001$ & $0.993 \pm 0.001$ \\
\hline \textbf{PriMIA} & $0.870 \pm 0.012$ & $0.860 \pm 0.002$ & $0.916 \pm 0.002$ \\
\hline \textbf{\ours (Ours)} & $0.980 \pm 0.004$ & $0.984 \pm 0.002$ & $0.984 \pm 0.002$ \\
\hline
\end{tabular}
    \caption{\textbf{The utilities of MLP classifiers for Pancreas cell classification task}{; mean $\pm$ SD.}}
    \label{tab:results_pancreas_mlp}
\end{table}

\begin{table}[h]
    \centering
    \begin{tabular}{|l|c|c|c|}
\hline & \textbf{Median F1 Scores} & \textbf{Weighted Precision} & \textbf{Weighted Recall }\\
\hline \textbf{Baron (P1)} & $0.982 \pm 0.003$ & $0.988 \pm 0.002$ & $0.987 \pm 0.002$ \\
\hline \textbf{Muraro (P2)} & $0.974 \pm 0.003$ & $0.966 \pm 0.002$ & $0.967 \pm 0.002$ \\
\hline \textbf{Seg (P3)} & $0.978 \pm 0.004$ & $0.986 \pm 0.004$ & $0.986 \pm 0.004$ \\
\hline \textbf{Wang (P4)} & $0.710 \pm 0.153$ & $0.840 \pm 0.048$ & $0.848 \pm 0.045$ \\
\hline \textbf{Xin (P5)} & $0.964 \pm 0.021$ & $0.977 \pm 0.006$ & $0.972 \pm 0.009$ \\
\hline \textbf{FL} & $0.989 \pm 0.002$ & $0.992 \pm 0.001$ & $0.992 \pm 0.001$ \\
\hline \textbf{PriMlA} & $0.935 \pm 0.004$ & $0.959 \pm 0.005$ & $0.936 \pm 0.020$ \\
\hline \textbf{\ours (Ours)} & $0.982 \pm 0.002$ & $0.987 \pm 0.002$ & $0.987 \pm 0.002$ \\
\hline
\end{tabular}
    \caption{\textbf{The utilities of SVCs for Pancreas cell classification task}{; mean $\pm$ SD.}}
    \label{tab:results_pancreas_svc}
\end{table}

\begin{table}[h]
    \centering
\begin{tabular}{|l|c|c|c|c|}
\hline & \textbf{Atelectasis} & \textbf{Effusion} & \textbf{Cardiomegaly} & \textbf{No Finding} \\
\hline \textbf{NIH (P1)} & $0.848 \pm 0.009$ & $0.938 \pm 0.003$ & $0.916 \pm 0.003$ & $0.917 \pm 0.004$ \\
\hline \textbf{PC (P2)} & $0.828 \pm 0.012$ & $0.932 \pm 0.002$ & $0.916 \pm 0.005$ & $0.912 \pm 0.002$ \\
\hline \textbf{CheX (P3)} & $0.848 \pm 0.005$ & $0.943 \pm 0.003$ & $0.900 \pm 0.008$ & $0.915 \pm 0.003$ \\
\hline \textbf{FL} & $0.909 \pm 0.003$ & $0.970 \pm 0.000$ & $0.958 \pm 0.002$ & $0.946 \pm 0.001$ \\
\hline \textbf{PriMIA} & $0.871 \pm 0.003$ & $0.958 \pm 0.001$ & $0.895 \pm 0.003$ & $0.931 \pm 0.001$ \\
\hline \textbf{\ours (Ours)} & $0.890 \pm 0.003$ & $0.963 \pm 0.001$ & $0.927 \pm 0.004$ & $0.936 \pm 0.002$ \\
\hline
\end{tabular}
    \caption{\textbf{The utilities of models pre-trained with MIMIC-XCR for pathology identification task}{; mean $\pm$ SD.}}
    \label{tab:results_xray_mimic}
\end{table}

\begin{table}[h]
    \centering
\begin{tabular}{|l|c|c|c|c|}
\hline & \textbf{Atelectasis} & \textbf{Effusion} & \textbf{Cardiomegaly} & \textbf{No Finding} \\
\hline \textbf{NIH (P1)} & $0.848 \pm 0.006$ & $0.938 \pm 0.004$ & $0.864 \pm 0.015$ & $0.915 \pm 0.003$ \\
\hline \textbf{PC (P2)} & $0.783 \pm 0.078$ & $0.904 \pm 0.049$ & $0.880 \pm 0.063$ & $0.880 \pm 0.062$ \\
\hline \textbf{CheX (P3)} & $0.836 \pm 0.007$ & $0.940 \pm 0.004$ & $0.897 \pm 0.007$ & $0.911 \pm 0.004$ \\
\hline \textbf{FL} & $0.893 \pm 0.006$ & $0.966 \pm 0.001$ & $0.953 \pm 0.003$ & $0.941 \pm 0.002$ \\
\hline \textbf{PriMIA} & $0.801 \pm 0.017$ & $0.893 \pm 0.021$ & $0.827 \pm 0.012$ & $0.878 \pm 0.012$ \\
\hline \textbf{\ours (Ours)} & $0.858 \pm 0.003$ & $0.934 \pm 0.005$ & $0.876 \pm 0.010$ & $0.908 \pm 0.005$ \\
\hline
\end{tabular}
    \caption{\textbf{The utilities of models pre-trained with ImageNet for pathology identification task}{; mean $\pm$ SD.}}
    \label{tab:results_xray_imgnet}
\end{table}

\begin{table}[h]
    \centering
    \begin{tabularx}{\textwidth}{|l|l|l|l|l|X|X|}
\hline & \textbf{Labels} & \textbf{Atelectasis} & \textbf{Effusion} & \textbf{Cardiomegaly} & \textbf{No Finding} &  \textbf{Total Number} \\
\hline
\multirow{2}{*}{ \textbf{NIH}} & 1 & 11559 & 13317 & 2776 & 60361 & 83519 \\
\cline { 2 - 6 } & 0 & 71960 & 70202 & 80743 & 23158 & \\
\hline \multirow{2}{*}{ \textbf{PC} } & 1 & 4345 & 3938 & 8744 & 48963 & 64143 \\
\cline { 2 - 6 } & 0 & 59798 & 60205 & 55399 & 15180 & \\
\hline \multirow{2}{*}{ \textbf{CheX} } & 1 & 29718 & 76894 & 23384 & 16970 & 120291 \\
\cline { 2 - 6 } & 0 & 17209 & 22035 & 18832 & 103321 & \\
\hline \multirow{2}{*}{ \textbf{MIMIC-CXR} } & 1 & 48790 & 57721 & 47673 & 81117 & 185452 \\
\cline { 2 - 6 } & 0 & 81723 & 93918 & 87561 & 104335 & \\
\hline
\end{tabularx}
    \caption{Chest radiology dataset information for training the models (with 4 outputs)}
    \label{tab:xray_dataset_stats}
\end{table}

\begin{table}[h]
    \centering
\begin{tabularx}{\textwidth}{|l|X|}
\hline \textbf{Name of the feature} & \textbf{Categories} \\
\hline
Mode of admission & Ground ambulance, No ambulance, Includes air ambulance, (no info) \\
\hline Triage level &  Resuscitation, Emergent, Urgent, Semi-urgent, Non-urgent, No information \\
\hline Readmission type & Planned readmission from previous acute care (no time restriction), Unplanned readmission within 7 days following discharge from acute care, Unplanned readmission 8 to 28 days following discharge from acute care, Unplanned readmission within 7 days following discharge from day surgery, New patient to the acute care unit, None of the above, no information \\
\hline Admission month & January to December (12 categories) \\
\hline Admission day (of the week) & Sunday to Saturday (7 categories) \\
\hline From acute care institute & Binary \\
\hline Gender & Binary (Male vs. Non-male) \\
\hline Receiving transfusion of plasma & Binary \\
\hline Receiving transfusion of platelet & Binary \\
\hline Receiving transfusion of Albumin & Binary \\
\hline Receiving X-ray of head & Binary \\
\hline Receiving Ultrasound of head & Binary \\
\hline Receiving X-ray of neck & Binary \\
\hline Receiving MRI of neck & Binary \\
\hline Receiving Ultrasound of neck & Binary \\
\hline Receiving MRI of chest & Binary \\
\hline Receiving Ultrasound of chest & Binary \\
\hline Receiving MRI of pelvis & Binary \\
\hline Receiving CT of limb & Binary \\
\hline Receiving MRI of limb & Binary \\
\hline Receiving X-ray of whole body & Binary \\
\hline Receiving CT of whole body & Binary \\
\hline Receiving MRI of whole body & Binary \\
\hline Receiving any test of shoulder & Binary \\
\hline Receiving any test of breast & Binary \\
\hline
\end{tabularx}
    \caption{Categorical Features of GEMINI dataset}
    \label{tab:cat_features_gemini}
\end{table}

\begin{table}[h]
    \centering
\begin{tabularx}{\textwidth}{|l|X|}
\hline \textbf{Name of the feature} & \textbf{Description} \\
\hline
Age & Patients' age \\
\hline Number of past visits &  Number of times a patient has visited hospitals in the past \\
\hline Blood transfusion of red blood cells &  Number of appropriate transfusions  \\
\hline CT test for head & Number of such test received by a patient  \\
\hline MRI test for head & Number of such test received by a patient \\
\hline CT test for neck & Number of such test received by a patient \\
\hline X-ray test for chest & Number of such test received by a patient \\
\hline CT test for chest & Number of such test received by a patient \\
\hline Echo test for chest & Number of such test received by a patient \\
\hline X-ray test for abdomen  & Number of such test received by a patient \\
\hline CT test for abdomen & Number of such test received by a patient \\
\hline MRI test for abdomen & Number of such test received by a patient \\
\hline Ultrasound test for abdomen & Number of such test received by a patient \\
\hline X-ray test for pelvis & Number of such test received by a patient \\
\hline CT test for pelvis & Number of such test received by a patient \\
\hline Ultrasound test for pelvis & Number of such test received by a patient \\
\hline X-ray test for limb & Number of such test received by a patient \\
\hline Ultrasound test for limb & Number of such test received by a patient \\
\hline Types of lab tests & The total number of unique test types received by the patient \\
\hline All available lab test results & The numerical results of 361 different types of lab tests. Each type of lab tests forms one feature. \\
\hline
\end{tabularx}
    \caption{Numerical Features of GEMINI dataset}
    \label{tab:num_features_gemini}
\end{table}

\clearpage
\bibliographystyle{vancouver}
\bibliography{my_bib}